\documentclass[letterpaper, 10 pt, conference]{ieeeconf}  %
\IEEEoverridecommandlockouts                              %

\usepackage{epsfig}
\usepackage[table]{xcolor}
\usepackage{graphicx}
\usepackage{amsmath}
\usepackage{amssymb}
\usepackage{mathtools}
\usepackage{color}
\usepackage{siunitx}
\usepackage{multirow}
\usepackage{wrapfig}
\usepackage{subcaption}
\usepackage{etoolbox}
\usepackage{xcolor}
\usepackage{booktabs}
\usepackage{cuted}
\usepackage{caption}
\usepackage{gensymb}

\newcommand{\kitti}[0]{KITTI~Scene~Flow }
\newcommand{\waymo}[0]{Waymo~Open~Dataset}

\newcommand{\flying}[0]{FlyingThings3D }

\newcommand{\lidar}[0]{LiDAR }
\newcommand{\transform}[0]{\mathbf{T}}

\newcommand{\transformAtT}[1]{\transform_{#1}}
\newcommand{\transformPrimeAtT}[1]{\transformAtT{#1}^*}
\newcommand{\pointcloud}[0]{\mathbf{P}}
\newcommand{\pointcloudAtT}[1]{\pointcloud_{#1}}
\newcommand{\point}[0]{\mathbf{p}}
\newcommand{\flowvec}[0]{\mathbf{f}}
\newcommand{\minflowspeed}[0]{f_{\text{min}}}
\newcommand{\vel}[0]{v}
\newcommand{\object}[0]{o}
\newcommand{\objects}[0]{\mathcal{O}}
\newcommand{\av}[0]{\text{AV}}
\newcommand{\tcurr}[0]{{t_{0}}}
\newcommand{\tprev}[0]{{t_{{\text -}1}}}
\newcommand{\modelname}[0]{FastFlow3D}
\newcommand{\modelabbreviation}[0]{FF3D}
\usepackage[pagebackref=true,breaklinks=true,letterpaper=true,colorlinks,bookmarks=false]{hyperref}

\newcommand\blfootnote[1]{%
  \begingroup
  \renewcommand\thefootnote{}\footnote{#1}%
  \addtocounter{footnote}{-1}%
  \endgroup
}

\newcommand{\secref}[1]{Section~\ref{#1}} %
\newcommand{\figref}[1]{Figure~\ref{#1}} %
\newcommand{\tabref}[1]{Table~\ref{#1}} %
\newcommand{\appref}[1]{Appendix~\ref{#1}} %

\newcommand{\laserfeature}[0]{l}

\newif\ificrafinal
\icrafinaltrue

\title{\LARGE \bf
Scalable Scene Flow from Point Clouds in the Real World
}

\author{Philipp Jund $^{*, 1}$, Chris Sweeney $^{*, 2}$, Nichola Abdo $^{2}$, Zhifeng Chen $^{1}$, Jonathon Shlens $^{1}$ \\ {\normalsize $^{1}$ Google Brain, $^{2}$ Waymo} \\ {\tt\small \{abdon,cjsweeney\}@waymo.com} }

\begin{document}

\maketitle
\thispagestyle{empty}
\pagestyle{empty}

\begin{abstract}
Autonomous vehicles operate in highly dynamic environments necessitating an accurate assessment of which aspects of a scene are moving and where they are moving to. A popular approach to 3D motion estimation, termed scene flow, is to employ 3D point cloud data from consecutive LiDAR scans, although such approaches have been limited by the small size of real-world, annotated LiDAR data.
In this work, we introduce a new large-scale dataset for scene flow estimation derived from corresponding tracked 3D objects, which is $\sim$1,000$\times$ larger than previous real-world datasets in terms of the number of annotated frames.
We demonstrate how previous works were bounded based on the amount of real LiDAR data available, suggesting that larger datasets are required to achieve state-of-the-art predictive performance.
Furthermore, we show how previous heuristics for operating on point clouds such as down-sampling heavily degrade performance, motivating a new class of models that are tractable on the full point cloud.
To address this issue, we introduce the \modelname~architecture which provides real time inference on the full point cloud. 
Additionally, we design human-interpretable metrics that better capture real world aspects by accounting for ego-motion and providing breakdowns per object type. 
We hope that this dataset may provide new opportunities for developing real world scene flow systems.
\end{abstract}

\section{Introduction}
\label{sec:intro}

\blfootnote{$^{*}$ Denotes equal contributions.}

Motion is a prominent cue that enables humans to navigate complex environments \cite{forsyth2002computer}. Likewise, understanding and predicting the 3D motion field of a scene -- termed the {\it scene flow} -- provides an important signal to enable autonomous vehicles (AVs) to understand and navigate highly dynamic environments \cite{thrun2006stanley}. Accurate scene flow prediction enables an AV to identify potential obstacles, estimate the trajectories of objects \cite{casas2018intentnet,chai2019multipath}, and aid downstream tasks such as detection, segmentation and tracking \cite{luo2018fast,mahjourian2018unsupervised}.

Recently, approaches that learn models to estimate scene flow from \lidar have demonstrated the potential for LiDAR-based motion estimation, outperforming camera-based methods \cite{liu2019flownet3d,wu2019pointpwc,gu2019hplflownet}. Such models take two consecutive point clouds as input and estimate the scene flow as a set of 3D vectors, which transform the points from the first point cloud to best match the second point cloud. One of the most prominent benefits of this approach is that it avoids the additional burden of estimating the depth of sensor readings as is required in camera-based approaches \cite{mahjourian2018unsupervised}.
Unfortunately, for LiDAR based data, ground truth motion vectors are ill-defined and not tenable because no correspondence exists between LiDAR returns from subsequent time points. Instead, one must rely on semi-supervised methods that employ auxiliary information to make strong inferences about the motion signal in order to bootstrap annotation labels \cite{menze2015object,geiger2012we}. 
Such an approach suffers from the fact that motion annotations are extremely limited (e.g. 400 frames in \cite{menze2015object,geiger2012we}) and often rely on pretraining a model based on synthetic data \cite{mayer2016flyingthings3d} which exhibit distinct noise and sensor properties from real data. 
Furthermore, previous datasets cover a smaller area, e.g., the KITTI scene flow dataset covers  1/5th the area of our proposed dataest. This allows for different subsampling tradeoffs and inspired a class of models that are not able to tractably scale training and inference beyond $\sim$10K points \cite{liu2019flownet3d,wu2019pointpwc,gu2019hplflownet,wang2020flownet3d++,liu2019meteornet}, making the usage of such models impractical in real world AV scenes which often contain 100K - 1000K points.

\begin{figure*}[t!]
    \centering 
    \setlength{\fboxsep}{0pt}
    \vspace{2mm}
    \fbox{\includegraphics[trim=0 110 0 0,clip, width=0.36\linewidth]{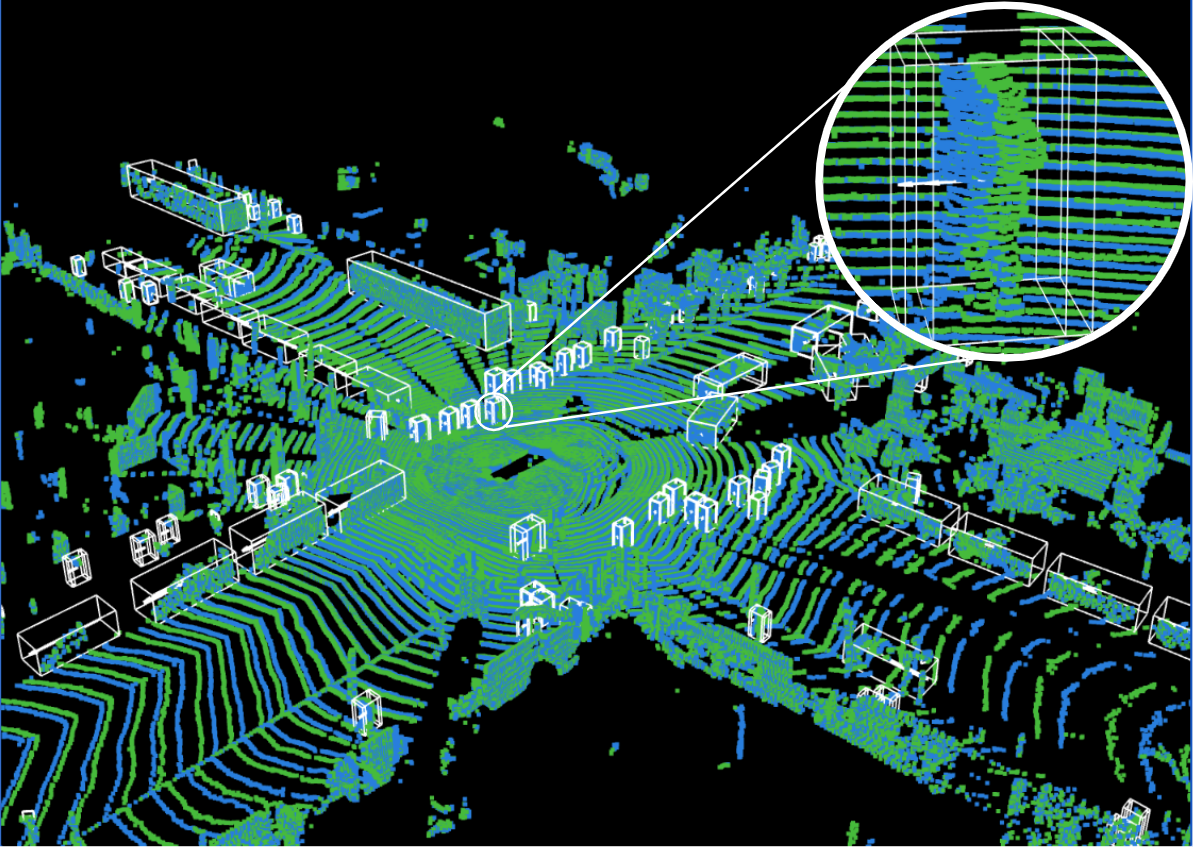}}
    \hspace{0.3cm}
    \fbox{\includegraphics[trim=0 110 0 0,clip,width=0.36\linewidth]{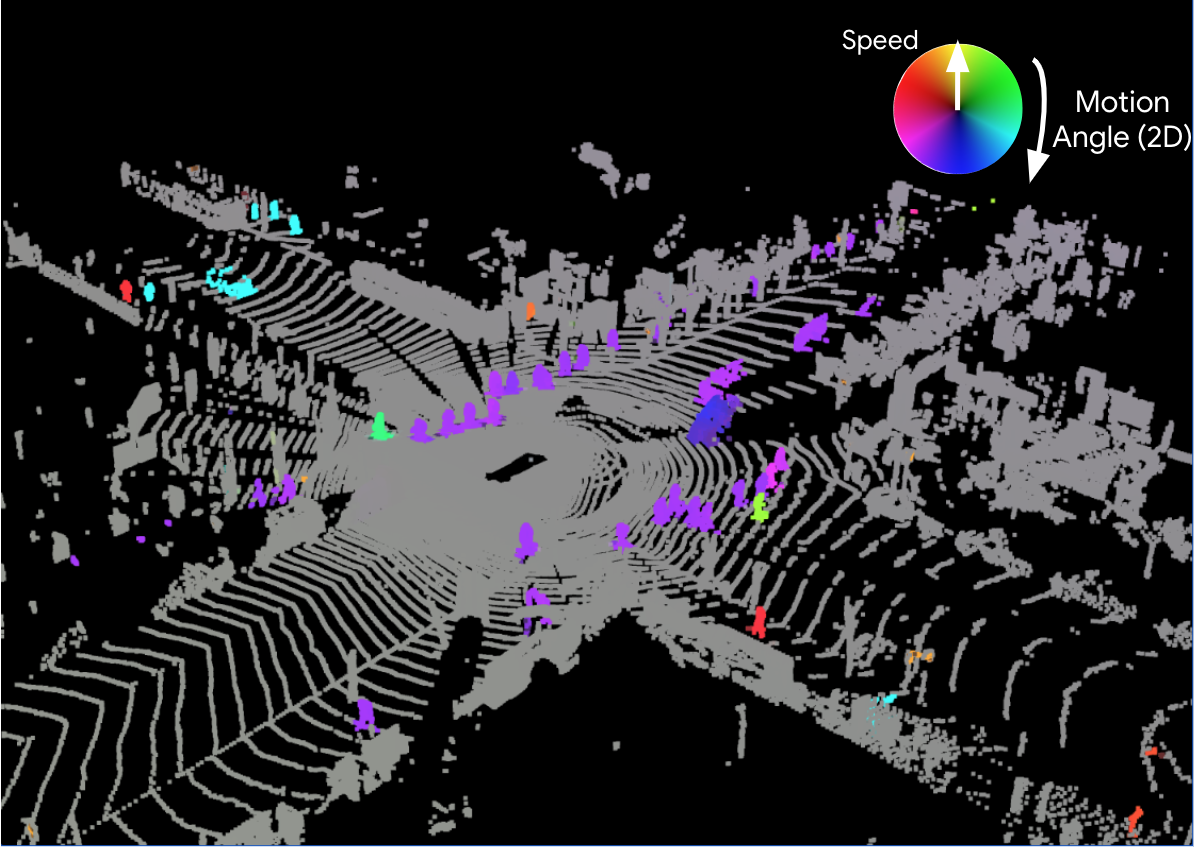}}
    \hspace{0.3cm}
    \includegraphics[width=0.178\linewidth]{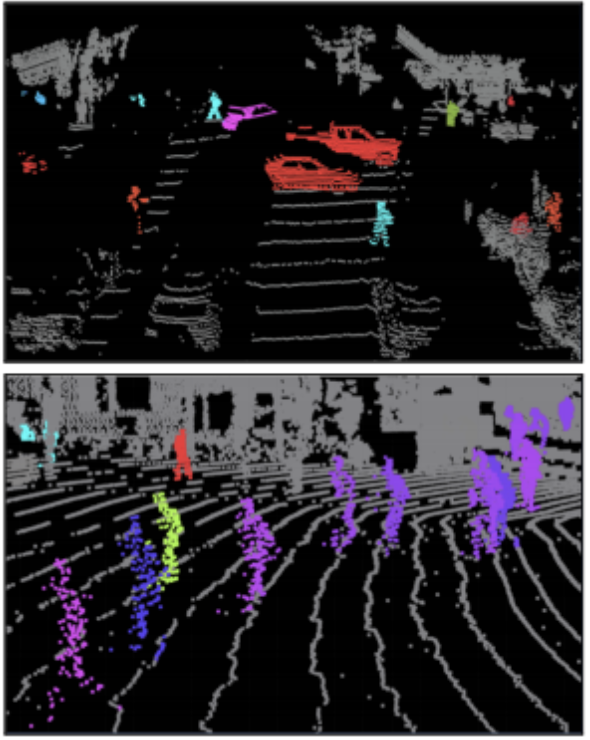}
    \captionof{figure}{\textbf{LiDAR scene flow estimation for autonomous vehicles.} Left: Overlay of two consecutive point clouds (green and blue, respectively) sampled at 10 Hz from the \waymo~\cite{sun2020scalability}. White boxes are tracked 3D bounding boxes for human annotated vehicles and pedestrians. Middle: Predicted scene flow for each point colored by direction, and brightened by speed based on overlaid frames$^\dagger$. Right: Two qualitative examples of bootstrapped annotations. \label{fig:motivation}}
\end{figure*}

In this work, we address these shortcomings of this field by introducing a large scale dataset for scene flow geared towards AVs. We derive per-point labels for motion estimation by bootstrapping from tracked objects densely annotated in a scene from a recently released large scale AV dataset \cite{sun2020scalability}.
The resulting scene flow dataset contains 198K frames of motion estimation annotations. This amounts to roughly $\sim$1,000$\times$ larger training set than the largest, commonly used real world dataset (200 frames) for scene flow \cite{menze2015object,geiger2012we}. By working with a large scale dataset for scene flow, we identify several indications that the problem is quite distinct from current approaches:
\begin{itemize}%
    \item Learned models for scene flow are heavily bounded by the amount of data. 
    \item Heuristics for operating on point clouds (e.g. downsampling) heavily degrade predictive performance. This observation motivates the development of a new class of models tractable on a full point cloud scene.
    \item Previous evaluation metrics ignore notable systematic biases across classes of objects %
    (e.g. predicting pedestrian versus vehicle motion).
\end{itemize}
We discuss each of these points in turn as we investigate working with this new dataset.
Recognizing the limitations of previous works, we develop a new baseline model architecture, {\it \modelname}, that is tractable on the complete point cloud with the ability to run in real time (i.e. $<$ 100 ms) on an AV. \figref{fig:motivation} shows scene flow predictions from \modelname,~trained on our scene flow dataset.
Finally, we identify and characterize an under-appreciated problem in semi-supervised learning based on the ability to predict the motion of unlabeled objects. We suspect the degree to which the fields of semi-supervised learning attack this problem may have strong implications for the real-world application of scene flow in AVs. We hope that the resulting dataset presented in this paper may open the opportunity for qualitatively new forms of learned scene flow models.

\section{Related Work}
\label{sec:related}

\subsection{Benchmarks for scene flow estimation}
Early datasets focused on the related problems of inferring depth from a single image \cite{saxena2006learning} or stereo pairs of images \cite{scharstein2002taxonomy,pfeiffer2013exploiting}. Previous datasets for estimating optical flow were small and largely based on synthetic imagery \cite{baker2011database,kondermann2012performance,morales2010ground,ladicky2012joint}.  Subsequent datasets focused on 2D motion estimation in movies or sequences of images \cite{butler2012naturalistic}. The KITTI Scene Flow dataset represented a huge step forward, providing
non-synthetic imagery paired with accurate ground truth estimates
However, it contained only 200 scenes for training and involved preprocessing steps that alter real-world characteristics~\cite{menze2015object}.  FlyingThings3D offered a modern large-scale synthetic dataset comprising $\sim$20K frames of high resolution data from which scene flow may be bootstrapped \cite{mayer2016flyingthings3d}. Internal datasets by~\cite{wang2018deep,lee2020pillarflow} are constructed similarly to ours, but are not publicly available and do not offer a detailed description. 
Recently, \cite{pontes2020scene} created two scene flow datasets in a similar fashion, subsampling 2,691 and 1,513 training scenes from the Argoverse~\cite{chang2019argoverse} and nuScenes~\cite{caesar2020nuscenes} datasets, resepectively. However, even without subsampling, larger datasets in terms of scenes and number of points are needed to train more accurate scene flow models as shown in~\cite{pontes2020scene} and Figure \ref{fig:dataset-size-study}.\blfootnote{$^\dagger$ Please note that we predict 3D flow, but color the direction of flow with respect to the $x$-$y$ plane for the visualization.}

\subsection{Datasets for tracking in AV's}
Recently, there have been several works introducing large-scale datasets for autonomous vehicle applications
\cite{geiger2012we,chang2019argoverse,caesar2020nuscenes,houston2020one,sun2020scalability}. While these datasets do not directly provide scene flow labels, they provide vehicle localization data, as well as raw \lidar data and bounding box annotations for perceived tracklets. These recent datasets offer an opportunity to propose a methodology to construct point-wise flow annotations from such data (\secref{sec:computing_groundtruth}).

We extend the~\waymo~(WOD) to construct a large-scale scene flow benchmark for dense point clouds~\cite{sun2020scalability}. We select the \waymo~because the bounding box annotations are at a higher acquisition frame (10 Hz) than competing datasets (e.g. 2 Hz in \cite{caesar2020nuscenes}) and contain $\sim$ $5\times$ the number of returns per \lidar frame (Table 1, \cite{sun2020scalability}).
In addition, the \waymo~also provides $\sim10\times$ more scenes and annotated \lidar frames than Argoverse \cite{chang2019argoverse}. Recently, \cite{houston2020one} released a large-scale dataset with 1,000+ hours of driving data. However, their tracked object annotations are not human-annotated but based on the results of the onboard perception system.

\subsection{Models for learning scene flow}

There is a rich literature of building learned models for scene flow using end-to-end learned architectures \cite{behl2019pointflownet,fan2019pointrnn,wu2019pointpwc,liu2019flownet3d,liu2019meteornet,wang2020flownet3d++,wu2020motionnet,lee2020pillarflow} as well as hybrid architectures \cite{dewan2016rigid,ushani2017learning,ushani2018feature}. We discuss these in Section \ref{sec:methods_model} in conjunction with building a scalable baseline model that operates in real time. Recently, Lee~\emph{et al.} presented an approach for predicting \emph{pillar-level} flow~\cite{lee2020pillarflow}. Whereas our model leverages a similar pillar-based architecture, we tackle the full scope of the scene flow problem and predict \emph{point-level} flow while being tractable enough for real-time applications.

Moreover, many previous works train models on synthetic datasets like FlyingThings3D \cite{mayer2016flyingthings3d} and evaluate and/or fine-tune on \kitti \cite{geiger2012we,menze2015object}. Typically, these models are limited in their ability to leverage synthetic data in training. This observation is in line with the robotics literature and highlights the challenges of generalization from the simulated to the real world \cite{bousmalis2018using,saxena2008robotic,viereck2017learning,gualtieri2016high}. 
\section{Constructing a Scene Flow Dataset}
In this section, we present an approach for generating scene flow annotations bootstrapped from existing labeled datasets. We first formalize the scene flow problem definition. We then detail our method for computing per-point flow vectors by leveraging the motion of 3D object label boxes. 
We emphasize that many details abound in the assumptions behind such annotations, how to calculate various transformations in the track labels, as well as how to handle important edge cases. 

\subsection{Problem definition}
\label{sec:prob_def}
We consider the problem of estimating 3D scene flow in settings where the scene at time $t_i$ is represented as a point cloud $\pointcloudAtT{i}$ as measured by a \lidar sensor mounted on the AV. Specifically, we define scene flow as the collection of 3D motion vectors $\flowvec~\coloneqq~(\vel^x, \vel^y, \vel^z)^\top$ for each point in the scene where $\vel^d$ is the velocity in the $d$ directions in $m/s$.

Following the scene flow literature, we predict flow given two consecutive point clouds of the scene, $\pointcloudAtT{-1}$ and $\pointcloudAtT{0}$. The scene flow encodes the motion between the previous and current time steps, $\tprev$ and $\tcurr$, respectively. 
We predict the scene flow at the current time step, $\pointcloudAtT{0}$ in order to make the predictions practical for real time operation. 

\subsection{From tracked boxes to flow annotations}
\label{sec:computing_groundtruth}

Obtaining ground truth scene flow from standard real-world \lidar data is a challenging task. One challenge is the lack of point-wise correspondences between subsequent \lidar frames. 
Manual annotation is too expensive and humans must contend with ambiguity due to changes in viewpoint and partial occlusions. Therefore, we focus on a scalable automated approach bootstrapped from existing labeled, tracked objects in \lidar data sequences.

The annotation procedure is straightforward. We assume that labeled objects are rigid and calculate point velocities using a {\it secant line approximation}. For each point $\point_{0}$ at time $\tcurr$, we compute the flow annotation as $\flowvec = \frac{1}{\Delta_t}(\point_{0} - \point_{-1})$, where $\Delta_t = \tcurr - \tprev$, $\point_{-1} = \transformAtT{\Delta}\,\point_{0}$ is the corresponding point at $\tprev$, and $\transformAtT{\Delta}$ is a homogeneous transformation inferred from the track labels of the object to which the point belongs (If there is no label at $\tprev$, the flow is annotated as invalid.). This captures how a moving object may have varying per-point flow magnitudes and directions. Though our rigidity assumption does not necessarily apply to non rigid objects (e.g. pedestrians), the high frame rate (10 Hz) minimizes non-rigid deformations between adjacent frames.

In order to calculate the transformation $\transformAtT{\Delta}$, we compensate for the ego motion of the \av~because this leads to superior predictive performance since a learned model does not need to additionally infer the AV's motion (most AVs are equipped with an IMU/GPS system to provide such information). Furthermore, compensating for ego motion improves the interpretability of the evaluation metrics (\secref{sec:methods_metrics}) since the predictions are now independent of the \av~motion.
We use this approach to compute the flow vectors for all points in $\pointcloudAtT{0}$ belonging to labeled objects. Points outside the labeled objects are assigned a flow of 0 $m/s$. This stationary assumption works well in practice, but has a notable gap when considering unlabeled moving objects in the scene. See Section \ref{sec:generalization} for an in depth analysis on how our model can generalize to unlabeled moving objects.

In this work, we apply this methodology on the~\waymo~\cite{sun2020scalability}. The dataset offers a large scale with diverse \lidar scenes where objects have been manually and accurately annotated with 3D boxes at 10Hz. Finally, the accurate \av~pose information permits compensating for ego motion. We note that the method for scene flow annotation is general and may be used to estimate 3D flow vectors of the label box poses available in other datasets \cite{caesar2020nuscenes,houston2020one,sun2020scalability,yu2020bdd100k}.

\section{Evaluation Metrics for Scene Flow}
\label{sec:methods_metrics}

Two common metrics used for 3D scene flow are mean $L_2$ error of pointwise flow and the percentage of predictions with $L_2$ error below a given threshold~\cite{liu2019flownet3d,wang2018deep}. In this work, we additionally propose modifications to improve the interpretability of the results.
\linebreak\linebreak
\noindent\textbf{Breakdown by object type.} Objects within the AV scene (e.g. vehicles, pedestrians) have different speed distributions dictated by the object class (Section \ref{sec:resultsDataset}). This becomes especially apparent after accounting for ego motion. Reporting a single error ignores these systematic differences.%
In practice, we find it more meaningful to report all prediction performances delineated by the object label.
\linebreak\linebreak
\textbf{Binary classification formulation.} One important practical application of predicting scene flow is enabling an \av~to distinguish between {\it moving} and {\it stationary} parts of the scene. In that spirit, we formulate a second set of metrics that represent a ``lower bar'' which captures a useful rudimentary signal. We employ this metric exclusively for the more difficult task of semi-supervised learning (Section \ref{sec:generalization}) where learning is more challenging. In particular, we assign a binary label to each reflection as either {\it moving} or {\it stationary} based on a threshold, $|\flowvec| \geq \minflowspeed$. 
Accordingly, we compute precision and recall metrics for these binary labels across an entire scene. Selecting a threshold, $\minflowspeed$, is not straightforward as there is an ambiguous range between very slow and stationary objects. For simplicity, we select a conservative threshold of $\minflowspeed =$ 0.5 m/s (1.1 mph) to assure that things labeled as moving are actually moving.

\section{\modelname: A Scalable Baseline Model}
\label{sec:methods_model}

\begin{figure}[t]
    \centering
    \vspace{3mm}
    \resizebox{0.5\textwidth}{!}{
    \Large
    \begin{picture}(623,275)
    \put(0,20){\includegraphics{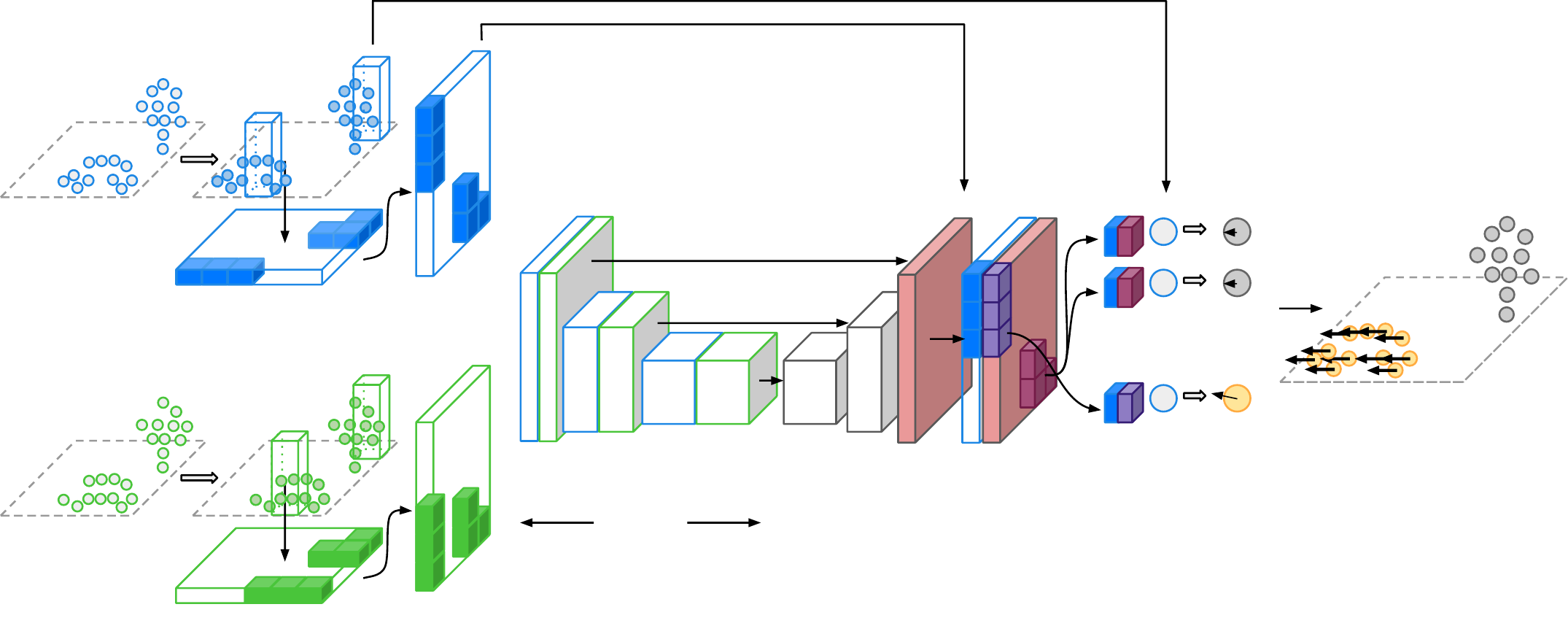}}
    \put(17,55){$t_{-1}$}
    \put(17,183){$t_{0}$}
    
    \put(205,211){\begin{tabular}[t]{c}\emph{concat}\\\emph{in depth}\end{tabular}}
    \put(228,72){\begin{tabular}[t]{c}\emph{shared}\\\emph{weights}\end{tabular}}
    
    \put(60,18){\begin{tabular}[t]{c}PointNet\\(dynamic voxelization)\end{tabular}}
    \put(220,18){\begin{tabular}[t]{c}Convolutional\\Autoencoder (U-Net)\end{tabular}}
    \put(355,18){\begin{tabular}[t]{c}Grid Flow\\Embedding\end{tabular}}
    \put(440,18){\begin{tabular}[t]{c}Multilayer\\Perceptron\end{tabular}}
    \end{picture}
    }
    
    \caption{\textbf{Diagram of \modelname\,model.} \modelname~consists of 3 stages employing a PointNet encoder with dynamic voxelization \cite{qi2017pointnet,zhou2019multiview}, a convolutional autoencoder \cite{ronneberger2015unet,lang2018pointpillars} with weights shared across two frames, and a shared MLP to regress an embedding on to a point-wise motion prediction. For details, see~\secref{sec:methods_model} and Appendix in \cite{jund2021scalable}.}
    \vspace{-0.5cm}
    \label{fig:model-diagram}
\end{figure}
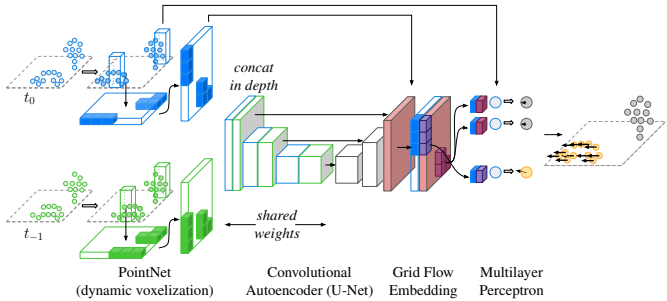

The average scene from the \waymo~consists of 177K points (Table \ref{table:dataset-comparison}), even though most models \cite{liu2019flownet3d,wu2019pointpwc,gu2019hplflownet,wang2020flownet3d++,liu2019meteornet} were designed to train with 8,192 points (16,384 points in \cite{liu2019meteornet}). This design choice favors algorithms that scale poorly to O(100K) regimes. For instance, many methods require preprocessing techniques such as nearest neighbor lookup. Even with efficient implementations \cite{zhou2008real,chen2019fast}, increasing fractions of inference time are dedicated to preprocessing instead of the core inference operation.
 
For this reason, we propose a new model that exhibits favorable scaling properties and may operate on O(100K) in a real time system. We name this model {\it \modelname} (\modelabbreviation). In particular we exploit the fact that LiDAR point clouds are dense, relatively flat along the $z$ dimension, but cover a large area along the $x$ and $y$ dimensions. The proposed model is composed of three parts: a scene encoder, a decoder fusing contextual information from both frames, and a subsequent decoder to obtain point-wise flow (Figure \ref{fig:model-diagram}).

\modelname~operates on two successive point clouds where the first cloud has been transformed into the coordinate frame of the second. The target annotations are correspondingly provided in the coordinate frame of the second frame. The result of these transformation is to remove apparent motion due to the movement of the \av~(\secref{sec:computing_groundtruth}). We train the resulting model with the average $L_2$ loss between the final prediction for each \lidar returns and the corresponding ground truth flow annotation \cite{wu2019pointpwc,liu2019flownet3d,gu2019hplflownet}.

The encoder computes embeddings at different spatial resolutions for both point clouds. The encoder is a variant of PointPillars \cite{lang2018pointpillars} and offers a great trade-off in terms of latency and accuracy by aggregating points within fixed vertical columns (i.e ``pillars'') followed by a 2D convolutional network to decrease the spatial resolution.
Each pillar center is parameterized through its center coordinate $(c_x, c_y, c_z)$. We compute the offset from the pillar center to the points in the pillar $(\Delta_x, \Delta_y, \Delta_z)$, and append the pillar center and laser features $(\laserfeature_0, \laserfeature_1)$, resulting in an 8D encoding $(c_x, c_y, c_z, \Delta_x, \Delta_y, \Delta_z, \laserfeature_0, \laserfeature_1)$.
Additionally, we employ dynamic voxelization \cite{zhou2019multiview}, computing a linear transformation and aggregating \emph{all} points within a pillar instead of sub-sampling points. Furthermore, we find that summing the featurized points in the pillar outperforms the max-pooling operation used in previous works~\cite{lang2018pointpillars,zhou2019multiview}.

One can draw an analogy of our pillar-based point featurization to more computationally expensive sampling techniques used by previous works \cite{liu2019flownet3d,wu2019pointpwc}. Instead of choosing representative sampled points based on expensive {\it farthest point sampling} and computing features relative to these points, we use a fixed grid to sample the points and compute features relative to each pillar in the grid. The pillar based representation allows our net to cover a larger area with an increased density of points. %

The decoder is a 2D convolutional U-Net \cite{ronneberger2015unet}. First, we concatenate the embeddings of both encoders at each spatial resolution. Subsequently, we use a 2D convolution to obtain contextual information at the different resolutions. These context embeddings are used as the skip connections for the U-Net, which progressively merges context from consecutive resolutions. To decrease latency, we introduce bottleneck convolutions and replace deconvolution operations (i.e. transposed convolutions) with bilinear upsampling \cite{odena2016deconvolution}. The resulting feature map of the U-Net decoder represents a grid-structured flow embedding. To obtain point-wise flow, we introduce the unpillar operation, which for each point retrieves the corresponding flow embedding grid cell, concatenates the point feature, and uses a multi layer perceptron to compute the flow vector.

As proof of concept, we showcase how the resulting architecture achieves favorable scaling behavior up to and beyond the number of laser returns in the \waymo~(Table \ref{table:latency-vs-points}). Note that we measure performance up to 1M points in order to accommodate multi-frame perception models which operate on point clouds from multiple time frames concatenated together \cite{ding20201st} \footnote{Many unpublished efforts employ multiple frames as detailed at \url{https://waymo.com/open/challenges}}. 
As mentioned earlier, previously proposed baseline models rely on nearest neighbor search for pre-processing, and even with an efficient implementation \cite{chen2019fast,zhou2008real} result in poor scaling behavior (see Section \ref{sec:results-baseline} for details.  ) making it prohibitively expensive to train and run these models on large, realistic datasets like the \waymo \footnote{In Section \ref{sec:results-baseline} we demonstrate that downsampling the point cloud severely degrades predictive performance further motivating architectures that can natively operate on the entire point cloud in real time.}.
In contrast, our baseline model exhibits nearly linear growth with a small constant. Furthermore, the typical period of a LiDAR scan is 10 Hz (i.e. 100 ms) and the latency of operating on 1M points is such that predictions may finish within the period of the scan as is required for real-time operation.

\begin{table}[t]
\vspace{2mm}
\normalsize
\rowcolors{2}{white}{gray!15}
    \centering
    \begin{tabular}{c|rrrr}
        \toprule
         & 32K & 100K & 255K & 1000K \\
        \midrule
        HPLFlowNet \cite{gu2019hplflownet}  & 431.1 & 1194.5 & {\footnotesize OOM} & {\footnotesize OOM} \\
        FlowNet3D \cite{liu2019flownet3d}  & 205.2 & 520.7 & 1116.4 & 3819.0 \\
        \modelname\,(ours) & 49.3 & 51.9 & 63.1  & 98.1 \\
        \bottomrule
    \end{tabular}
    \caption{\textbf{Inference latency varying point cloud sizes.} All numbers report latency in ms on a NVIDIA Tesla P100 with batch size $=$ 1. The timings for HPLFlowNet \cite{gu2019hplflownet} differ from reported results as we include
    the required preprocessing on the raw point clouds. {\footnotesize OOM} indicates out of memory.}
    \label{table:latency-vs-points}
    \vspace{-0.5cm}
\end{table}
\section{Results}
\label{sec:results}

We first present results describing the generated scene flow dataset and discuss how it compares to established baselines for scene flow in the literature (\secref{sec:resultsDataset}). In the process, we discuss dataset statistics and how this affects our selection of evaluation metrics. Next, in \secref{sec:results-baseline} we present the \modelname\,baseline architecture trained on the resulting dataset. We showcase with this model the necessity of training with the full density of point cloud returns as well as the complete dataset. These results highlight deficiencies in previous approaches which employed too few data or employed sub-sampled points for real-time inference. Finally, in \secref{sec:generalization} we discuss an extension to this work in which we examine the generalization power of the model and highlight an open challenge in the application of self-supervised and semi-supervised learning techniques.
\blfootnote{$^\ddagger$ indicates that only 400 frames of the KITTI dataset were annotated for scene flow (200 available for training). $^\dagger$ indicates the average number of points with distance from the camera $\leq$ 35.}

\subsection{A large-scale dataset for scene flow}
\label{sec:resultsDataset}

\begin{table}[t!]
\footnotesize    
\vspace{2mm}
\rowcolors{2}{gray!15}{white}
\begin{center}
\begin{tabular}{lccc}
\toprule
& KITTI&  \flying & Ours \\
\midrule
Data & LiDAR & Synth. & LiDAR \\
Label & Semi-Sup. & Truth & Super. \\

\midrule
Scenes & 22 & -- & 1150 \\ 
\# LiDAR Frames & 200 $^\ddagger$ & 28K & 198K \\
Avg Points/Frame & 208K & 220K $^\dagger$ & 177K \\
\bottomrule
\end{tabular}
\end{center}
\vspace{-0.2cm}
\caption{\textbf{Comparison of popular datasets for scene flow estimation}. \cite{geiger2012we} is computed through a semi-supervised procedure \cite{menze2015object}. \cite{mayer2016flyingthings3d} is computed from a depth map based on a geometric procedure \cite{liu2019flownet3d}. \# LiDAR frames counts annotated LiDAR frames for training and validation.}
\label{table:dataset-comparison}
\vspace{-0.3cm}
\end{table}

\begin{figure}[t]
\centering
\vspace{2mm}
\scriptsize
\rowcolors{2}{white}{gray!15}
\begin{tabular}{c|lrlr}
    \toprule
    & \multicolumn{2}{c}{moving} & \multicolumn{2}{c}{stationary} \\
    \midrule
     vehicles &  32.0\% & {\footnotesize (843.5M)} &  68.0\% & {\footnotesize (1,790.0M)} \\
     pedestrians & 73.7\% & {\footnotesize (146.9M)}  & 26.3\% & {\footnotesize (52.4M)} \\
     cyclists & 84.7\% & {\footnotesize (7.0M)} & 15.2\% & {\footnotesize (1.6M)} \\
     \bottomrule
\end{tabular}
\newline\newline \\
\includegraphics[width=0.7\linewidth]{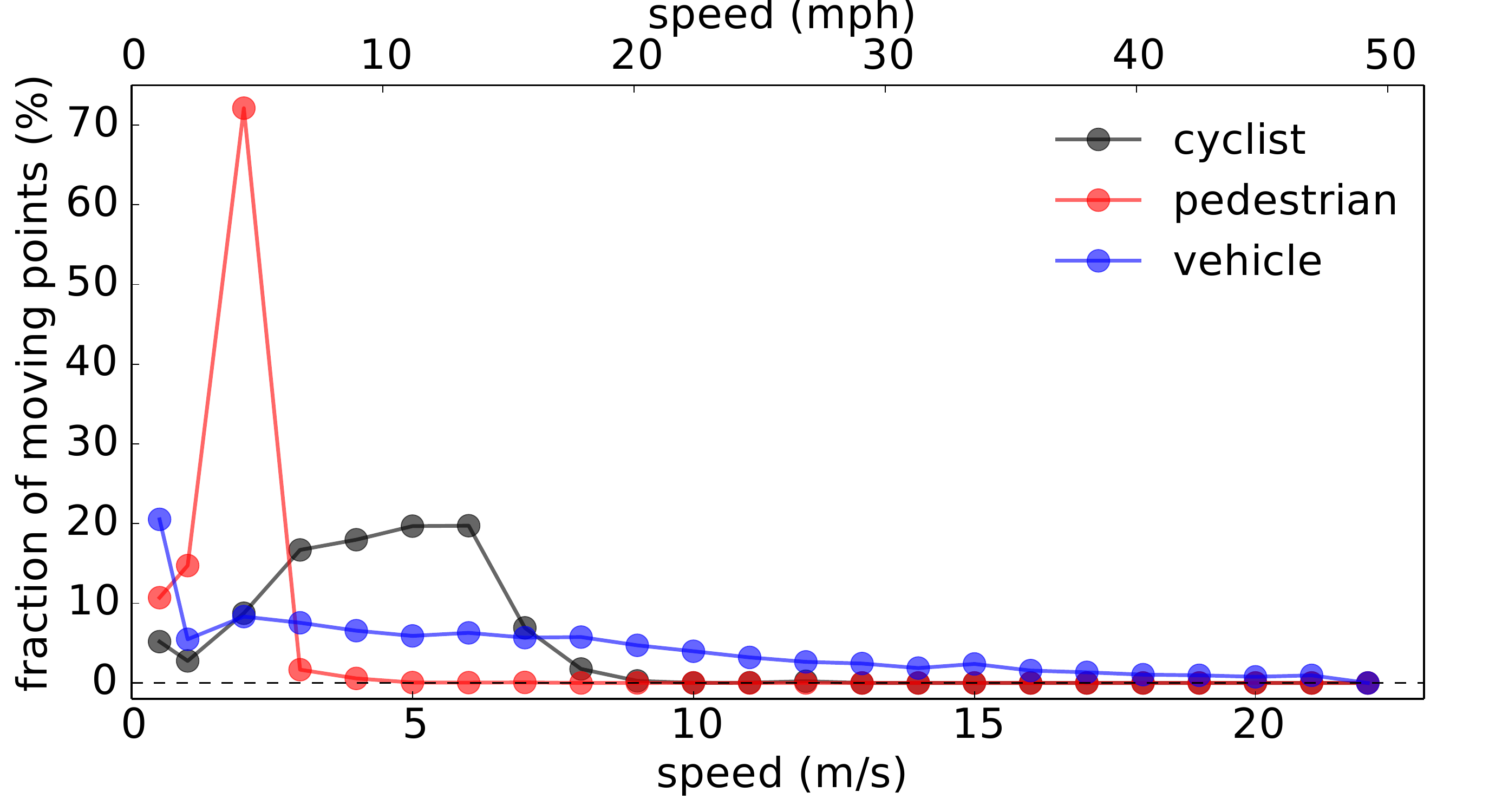}
\caption{\textbf{Distribution of moving and stationary~\lidar points}. Statistics computed from training set split. Top: Distribution of moving and stationary points across all frames (raw counts in parenthesis). We consider points with a flow magnitude below 0.1 m/s to be stationary. Bottom: Distribution of speeds for moving points.}
\label{fig:data-stats}
\vspace{-0.3cm}
\end{figure}

The \waymo~provides an accurate source of tracked 3D objects and an opportunity for deriving a large-scale scene flow dataset across a diverse and rich domain~\cite{sun2020scalability}. As previously discussed, scene flow ground truth does not exist in real-world point cloud datasets based on standard time-of-flight \lidar~because no correspondences exist between points from subsequent frames.

To generate a reasonable set of scene flow labels, we leveraged the human annotated tracked 3D objects from the \waymo~\cite{sun2020scalability}. Following the methodology in \secref{sec:computing_groundtruth}, we derived a supervised label $(\vel^x, \vel^y, \vel^z)$ for each point in the scene across time.
\figref{fig:motivation} (right) highlights some qualitative examples of the resulting annotation of scene flow using this methodology. In the selected frames, we highlight the diversity of the scene and difficulty of the resulting bootstrapped annotations. Namely, we observe the challenges of working with real \lidar data including the noise inherent in the sensor reading, the prevalence of occlusions and variation in object speed. All of these qualities result in a challenging predictive task.

The dataset comprises 800 and 200 scenes, termed {\it run segments}, for training and validation, respectively. Each run segment is 20 seconds recorded at 10 Hz \cite{sun2020scalability}. Hence, the training and validation splits contain 158,081 and 39,987 frames.\footnote{Please see the Appendix in \cite{jund2021scalable} for more details on downloading and accessing this new dataset.} The total dataset comprises 24.3B and 6.1B \lidar returns in each split, respectively. Table \ref{table:dataset-comparison} indicates that the resulting dataset is orders of magnitude larger than the standard KITTI scene flow dataset \cite{geiger2012we,menze2015object} and even surpasses the large-scale synthetic dataset \flying  \cite{mayer2016flyingthings3d} often used for pretraining.

\figref{fig:data-stats} provides a summary of the scene flow constructed from the \waymo. Across 7,029,178 objects labeled across all frames \footnote{A single instance of an object may be tracked across $N$ frames. We count a single instance as $N$ labeled objects.}, we find that $\sim$64.8\% of the points within pedestrians, cyclists and vehicles are stationary. This summary statistic belies a large amount of systematic variability across object class. For instance, the majority of points within vehicles (68.0\%) are parked and stationary, whereas the majority of points within pedestrians (73.7\%) and cyclists (84.7\%) are actively moving. The motion signature of each class of labeled object becomes even more distinct when examining the distribution of moving objects (\figref{fig:data-stats}, bottom). Note that the average speed of moving points corresponding to pedestrians (1.3 m/s or 2.9 mph), cyclists (3.8 m/s or 8.5 mph) and vehicles (5.6 m/s or 12.5 mph) vary  significantly. This variability of motion across object types emphasizes our selection of evaluation metrics that consider the prediction of each class separately.

\subsection{A scalable model baseline for scene flow}
\label{sec:results-baseline}

\begin{figure}[t!]
    \centering
    \vspace{2mm}
    \includegraphics[width=0.44\linewidth]{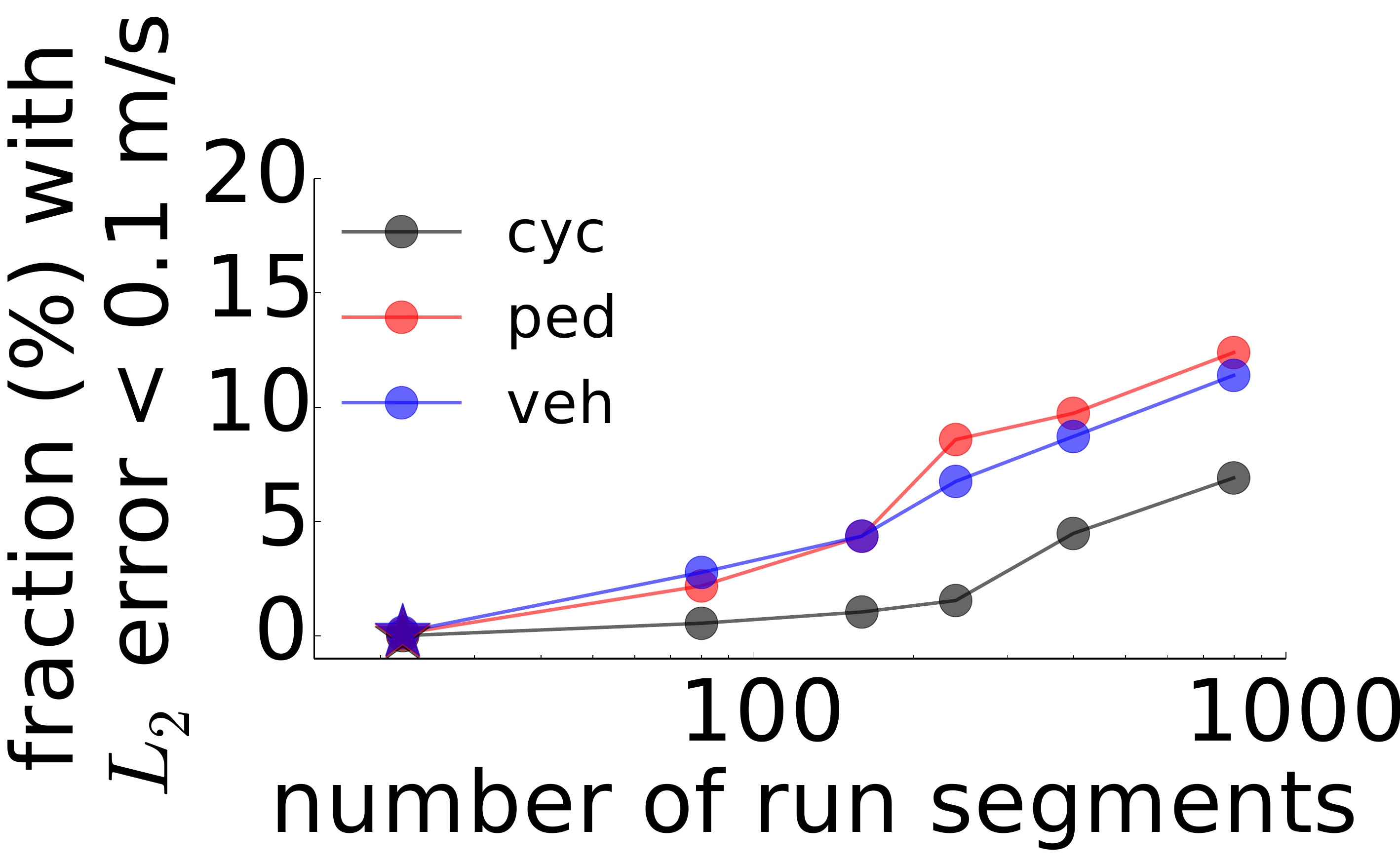}
    \includegraphics[width=0.44\linewidth]{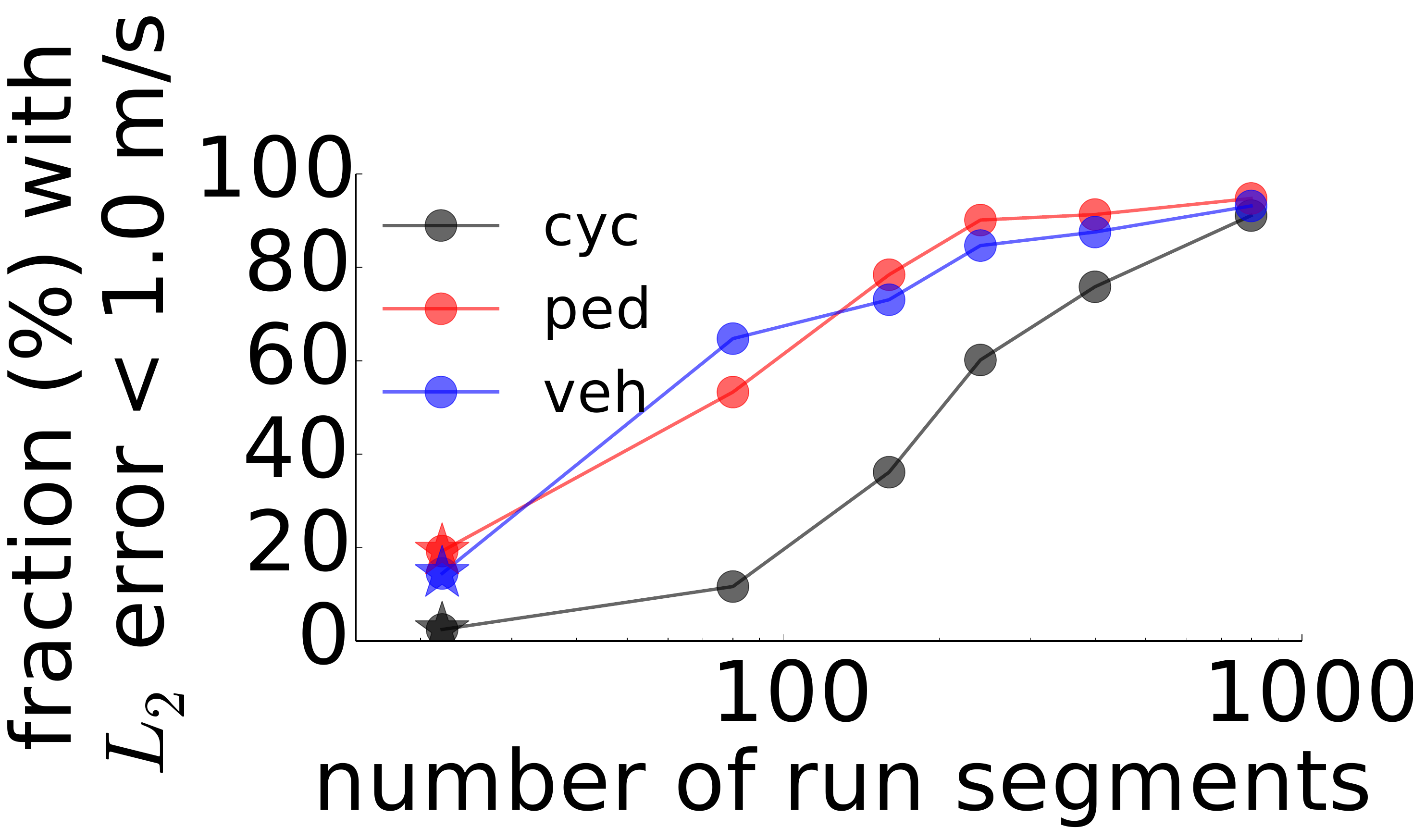}
    \caption{\textbf{Accuracy of scene flow estimation is bounded by the amount of data.} Each point corresponds to the cross validated accuracy of a model trained on increasing amounts of data (see text). $Y$-axis reports the fraction of \lidar~returns contained within {\it moving} objects whose motion vector is  estimated within 0.1 m/s (top) or 1.0 m/s (bottom) $L_2$ error. Higher numbers are better. The star indicates a model trained on the number of run segments in \cite{geiger2012we,menze2015object}.}
    \label{fig:dataset-size-study}
    \vspace{-0.55cm}
\end{figure}

\begin{figure}[t!]
    \centering    
    \vspace{2mm}
    \includegraphics[width=0.44\linewidth]{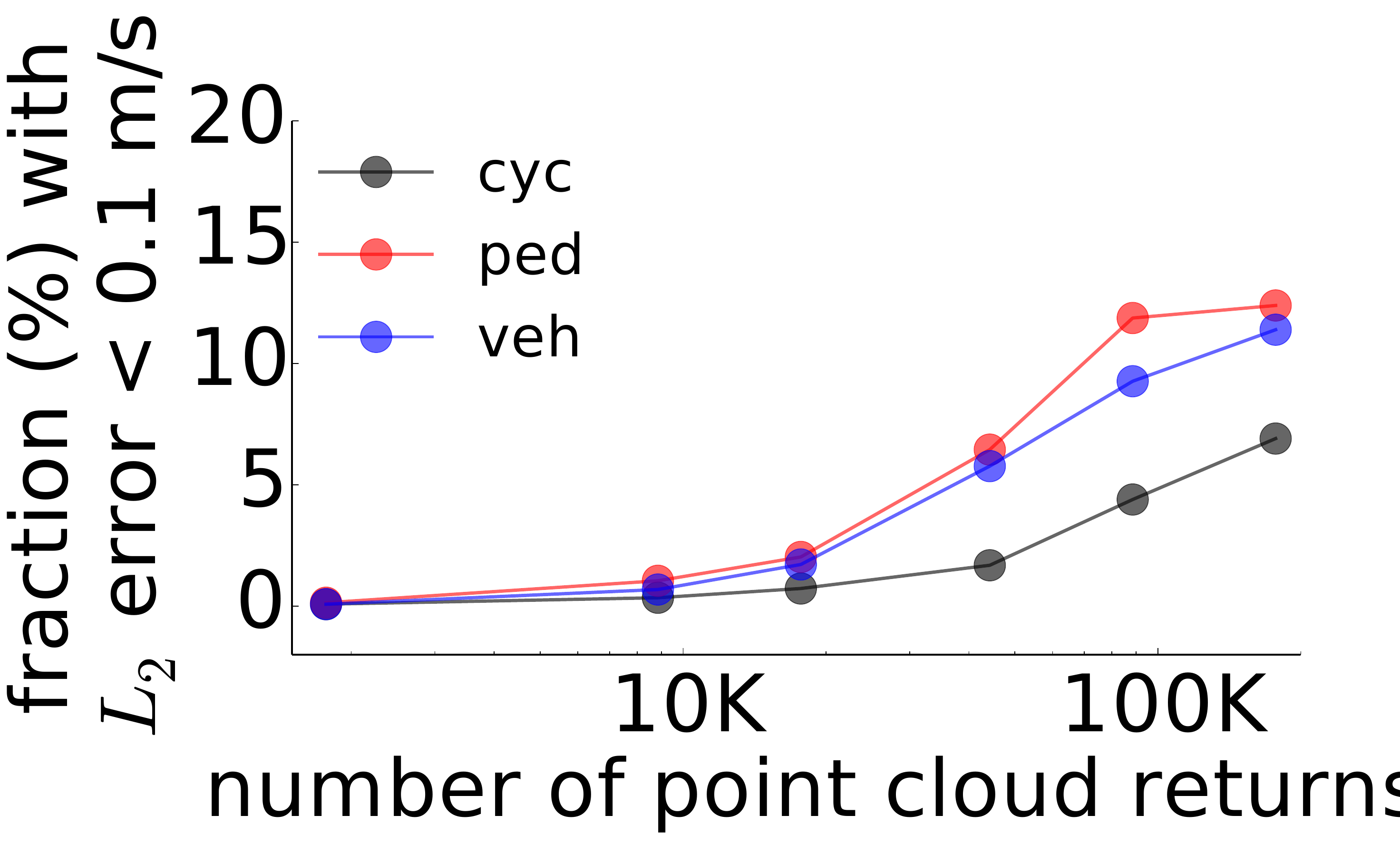}
    \includegraphics[width=0.44\linewidth]{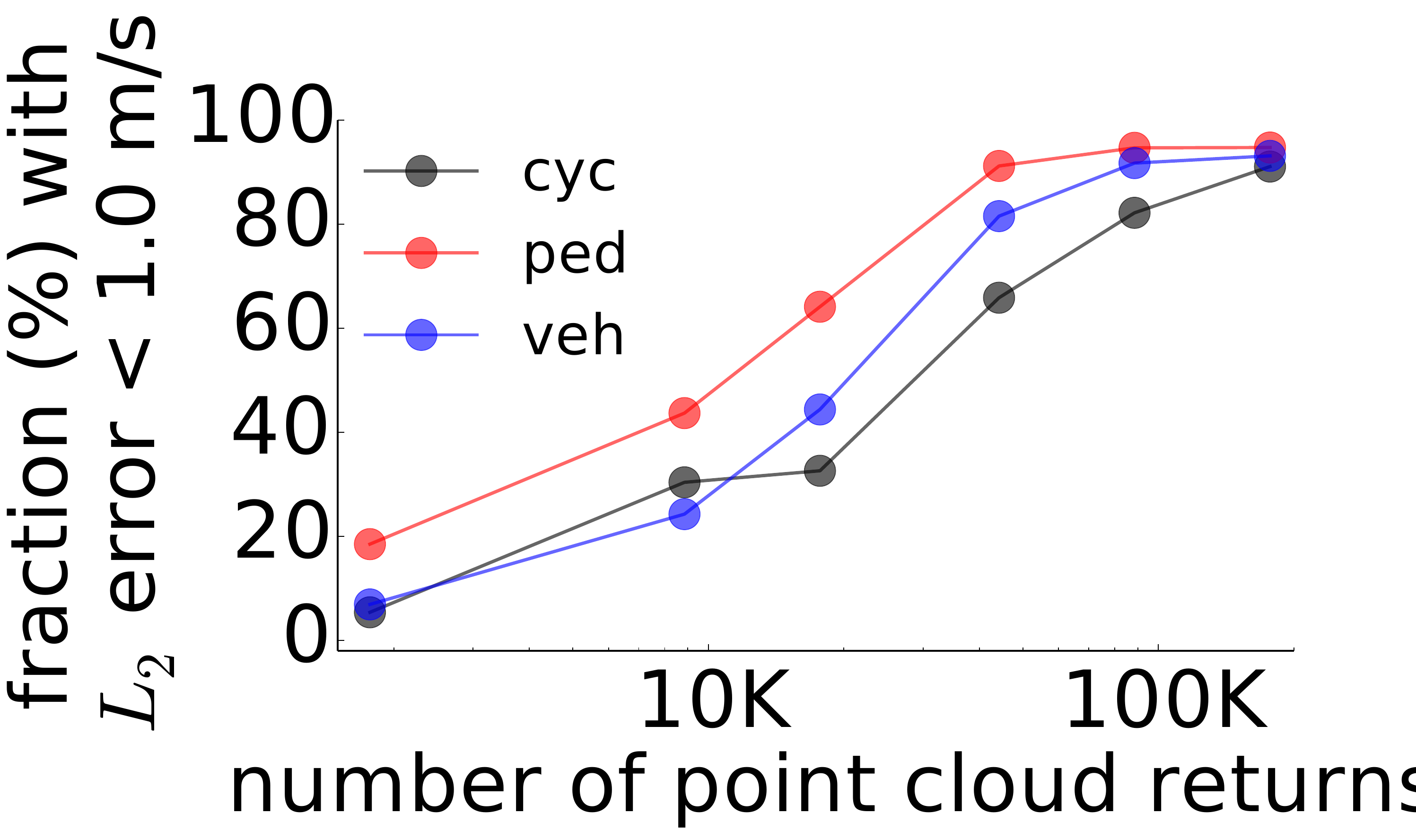}
    \caption{\textbf{Accuracy of scene flow estimation requires the full density of the point cloud scene.} Each point corresponds to the cross validated accuracy of a model trained on an increasing density of point cloud points. $Y$-axis reports the fraction of \lidar~returns contained within {\it moving} vehicles, pedestrians and cyclists whose motion vector is correctly estimated within 0.1 m/s (top) and 1.0 m/s (bottom) $L_2$ error.}
    \label{fig:density-study}
    \vspace{-0.6cm}
\end{figure}

We train the \modelname\, architecture on the scene flow data. Briefly, the architecture consists of 3 stages employing established techniques: (1) a PointNet encoder with a dynamic voxelization \cite{zhou2019multiview,qi2017pointnet}, (2) a convolutional autoencoder with skip connections \cite{ronneberger2015unet} in which the first half of the architecture \cite{lang2018pointpillars} consists of shared weights across two frames, and (3) a shared MLP to regress an embedding on to a point-wise motion prediction.

The resulting model contains 5.23M parameters, a vast majority of which reside in the convolution architecture (4.21M). A small number of parameters (544) are dedicated to featurizing each point cloud point \cite{qi2017pointnet} as well as performing the final regression on to the motion flow (4,483). These latter sets of parameters are purposefully small in order to effectively constrain computational cost because they are applied across all $N$ points in a point cloud.

We evaluate the resulting model on the cross-validated split using the aforementioned metrics across an array of experimental studies to justify the motivation for this dataset as well as demonstrate the difficulty of the prediction task.

\begin{table*}[t]
\footnotesize
    \centering
            \vspace{2mm}

    \rowcolors{3}{gray!15}{white}
     \bgroup
     \def\arraystretch{1.02}%
    \begin{tabular}{c|ccc|ccc|ccc|c}
        \toprule
        \multirow{2}{*}{error metric} &
        \multicolumn{3}{c|}{vehicle} & \multicolumn{3}{c|}{pedestrian} & \multicolumn{3}{c|}{cyclist} & \multirow{2}{*}{background} \\
        & all & moving & stationary & all & moving & stationary & all & moving & stationary & \\
        \midrule
        mean (m/s)\,\, & 0.18 & 0.54 & 0.05  &  0.25 & 0.32 & 0.10 & 0.51 & 0.57 & 0.10 & 0.07 \\
        mean (mph) & 0.40&  1.21&  0.11 &  0.55 &  0.72 & 0.22 &  1.14&  1.28&  0.22 &  0.16 \\
        \midrule
        $\leq$ 0.1 m/s & 70.0\% & 11.6\%& 90.2\%& 33.0\%& 14.0\%& 71.4\% & 13.4\%& 4.8\% & 78.0\% & 95.7\% \\
        $\leq$ 1.0 m/s & 97.7\%& 92.8\%& 99.4\% & 96.7\%& 95.4\%& 99.4\% & 89.5\%& 88.2\%& 99.6\%& 96.7\%\\
        \bottomrule
    \end{tabular}
    \egroup
    \caption{\textbf{Performance of baseline on scene flow in large-scale dataset}. Mean pointwise $L_2$ error (top) and percentage of points with error below 0.1 m/s and 1.0 m/s (bottom). Most errors are $\leq$ 1.0 m/s. Additionally, we investigate the error for stationary and moving points where a point is coarsely considered moving if the flow vector magnitude is $\geq0.5$~m/s.}
    \label{table:central-results}
    \vspace{-0.2cm}
\end{table*}

We first approach the question of what the appropriate dataset size is given the prediction task. \figref{fig:dataset-size-study} provides an ablation study in which we systematically subsample the number of run segments employed for training \footnote{We subsample the number of run segments and not frames because subsequent frames within a single run segment are heavily correlated.}.
We observe that predictive performance improves significantly as the model is trained on increasing numbers of run segments. We find that cyclists trace out a curve quite distinct from pedestrians and vehicles, possibly indicative of the small number of cyclists in a scene (\figref{fig:data-stats}). Secondly, we observe that the cross validated accuracy is far from saturating behavior when approximating the amount of data available in the KITTI scene flow dataset \cite{geiger2012we,menze2015object} (\figref{fig:dataset-size-study}, stars). We observe that even with the complete dataset, our metrics do not appear to exhibit asymptotic behavior indicating that models trained on the \waymo~may still be data bound. This result parallels detection performance reported in the original results (Table 10 in \cite{sun2020scalability}). 

We next investigate how scene flow prediction is affected by the density of the point cloud scene. This question is important because many baseline models purposefully operate on a smaller number of points (\tabref{table:latency-vs-points}) and by necessity must heavily sub-sample the number of points in order to perform inference in real time. In stationary objects, we observe minimal detriment in performance (data not shown). This result is not surprising given that the vast majority of LiDAR returns arise from stationary, background objects (e.g. buildings, roads).
However, we do observe that training on sparse versions of the original point cloud severely degrades predictive performance of {\it moving} objects (\figref{fig:density-study}). Notably, moving pedestrians and vehicle performance appear to be saturating indicating that if additional LiDAR returns were available, they would have minimal additional benefit in terms of predictive performance. 

In addition to decreasing point density, previous works also filter out the numerous returns from the ground in order to limit the number of points to predict \cite{wu2019pointpwc,liu2019flownet3d,gu2019hplflownet}. Such a technique has a side benefit of bridging the domain gap between \flying and KITTI~Scene~Flow, which differ in the inclusion of such points. We performed an ablation experiment to parallel this heuristic by training and evaluating with our annotations but removing points with a crude threshold of 0.2 m above ground. When removing ground points, we found that the mean $L_2$ error increased by 159\% and 31\% for points in moving and stationary objects, respectively. We take these results to indicate that the inclusion of ground points provide a useful signal for predicting scene flow.
Taken together, these results provide post-hoc justification for building an architecture which may be tractably trained on all point cloud returns instead of one that only trains on a sample of the returns.

Finally, we report our results on the complete dataset and identify systematic differences across object class and whether or not an object is moving~(\tabref{table:central-results}). Producing baseline comparisons for previous nearest neighbor based models is prohibitively expensive due to their poor scaling behavior. \footnote{ We did try experiments involving cropping and downsampling the point clouds to make comparison to the baselines feasible. However, these modifications distorted the points too much to serve as practical input data.} We hope to motivate a new class of real time scene flow models that are capable of training on our dataset.

\tabref{table:central-results} indicates that {\it moving} vehicle points have a mean $L_2$ error of 0.54~m/s, corresponding to 10\% of the average speed of moving vehicles (5.6 m/s). Likewise, the mean $L_2$ error of moving pedestrian and cyclist points are 0.32~m/s and 0.57~m/s, corresponding to 25\% and 15\% of the mean speed of each object class, respectively. Hence, the ability to predict vehicle speed is better than pedestrians and cyclists.
We suspect that these imbalances are largely due to imbalances in the number of training examples for each label and the average speed of these objects. For instance, the vast majority of points are marked as {\it background} and hence have a target of zero motion. Because the background points are dominant, we likewise observe the error to be smallest.
\subsection{Generalizing to unlabeled moving objects}
\label{sec:generalization}
\begin{figure*}[t]
    \centering
    \setlength{\fboxsep}{0pt}
    \fbox{\includegraphics[width=0.9\linewidth]{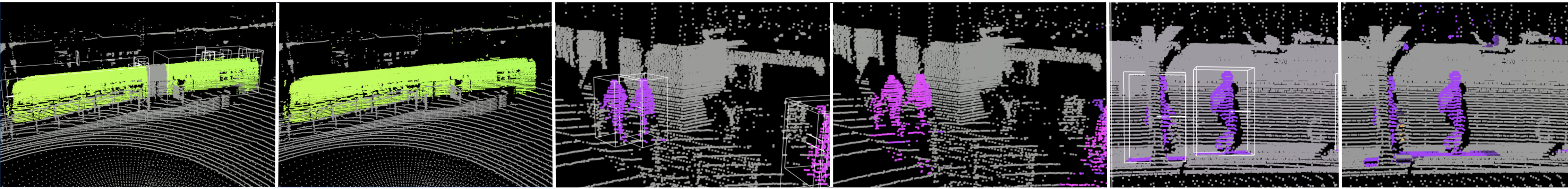}}
    \caption{\textbf{Generalizing to unlabeled moving objects.} Three examples each with the bootstrapped annotation (left) and model prediction (right) (Color code from Figure \ref{fig:motivation}). Left example: Despite missing flow annotation for the middle of a bus, our model can generalize well. Middle example: The model generalizes an unlabeled object (moving shopping cart). Right example: failures of generalization as motion is incorrectly predicted for the ground and parts of the tree.}
    \label{fig:qualitative-generalization}
    \vspace{-0.2cm}
\end{figure*}

The mean $L_2$ error is averaged over many points, making it unclear if this statistic may be dominated by outlier events. To address this issue, we show the percentage of points in the \waymo~evaluation set with $L_2$ errors below 0.1 m/s and 1.0 m/s. We observe that the vast majority of the errors are below 1.0 m/s (2.2 mph) in magnitude, indicating a rather regular distribution to the residuals.
For example, the residuals of 92.8\% and 99.8\% of moving and stationary vehicle points have an error below 1.0 m/s. %
In the next section, we also investigate how the prediction accuracy for classes like pedestrians and cyclists can be cast as a discrete task distinguishing moving and stationary points.

\begin{table}[t]
\scriptsize
    \centering
    \rowcolors{3}{gray!15}{white}
    \begin{tabular}{c|c|ccc|cc}
        \toprule
        & \multirow{2}{*}{method} & \multicolumn{3}{c|}{$L_2$ error (m/s)} & \multirow{2}{*}{prec} & \multirow{2}{*}{recall} \\
        & & all & moving & stationary & & \\
        \midrule
        & supervised & 0.51 & 0.57 & 0.10 & 1.00 & 0.95 \\
        cyc & stationary & 1.13 & 1.24 & 0.06 & 1.00 & 0.67 \\
        & ignored & 0.83 & 0.93 & 0.06 & 1.00 & 0.78 \\
        \midrule
        & supervised & 0.25 & 0.32 & 0.10 & 1.00 & 0.91 \\
        ped & stationary & 0.90 &1.30 & 0.10 & 0.97 & 0.02\\
        & ignored &  0.88 & 1.25 & 0.10 & 0.99 & 0.07 \\
        \bottomrule
    \end{tabular}
    \caption{\textbf{Generalization of motion estimation.} Approximating generalization for moving objects by artificially excluding a class from training by either treating all its points as having zero flow (stationary) or as having no target label (ignored). We report the mean pointwise $L_2$ error and the precision and recall for moving point classification.}
    \label{table:generalization}
    \vspace{-0.45cm}
\end{table}

Our supervised method for generating flow ground truth relies on every moving object having an accompanying tracked box. Without a tracked box, we effectively assume the points on an object are stationary. Though this assumption holds for the vast majority of points, there are still a wide range of moving objects that our algorithm assumes to be stationary. For deployment on a safety critical system, it is important to capture motion for these objects (e.g. stroller, opening car doors, etc.). Even though the labeled data does not capture such objects, we find qualitatively that a trained model does capture some motion in these objects (Figure \ref{fig:qualitative-generalization}). We next ask the degree to which a model trained on such data predicts the motion of unlabeled moving objects.

To answer this question, we construct several experiments by artificially removing labeled objects from the scene and measuring the ability of the model (in terms of the point-wise mean $L_2$ error) to predict motion in spite of this disadvantage. Additionally, we coarsely label points as {\it moving} if their annotated speed (flow vector magnitude) is $\geq$ 0.5 m/s ($\minflowspeed$) and query the model to quantify the precision and recall for moving classification. This latter measurement of {\it detecting} moving objects is particularly important for guiding planning in an AV \cite{mcnaughton2011motion,chu2012local,dolgov2008practical}.

\tabref{table:generalization} reports results for selectively ablating the labels for pedestrian and cyclist. We ablate the labels in two methods: (1) {\it Stationary} treats points of ablated objects as background with no motion, (2) {\it Ignored} treats points of ablated objects as having no target label.
We observe that {\it fixing} all points as  stationary results in a model with near perfect precision. However, the recall suffers enormously, particularly for pedestrians. Our results imply that unlabeled points predicted to be {\it moving} are almost perfectly correct (i.e. minimal false positives), however the recall is quite poor as many moving points are not identified (i.e. large number of false negatives). We find that treating the unlabeled points as {\it ignored} improves the performance slightly, indicating that even moderate information known about potential moving objects may alleviate challenges in recall.
Notably, we observe a large discrepancy in recall between the ablation experiments for cyclists and pedestrians. We posit that this discrepancy is likely due to the much larger amount of pedestrian labels in the \waymo. Removing the entire class of pedestrian labels removes much more of the ground truth labels for moving objects.

Although our model has some capacity to generalize to unlabeled moving object points, this capacity is limited. Ignoring labeled points does mitigate the error rate for cyclists and pedestrians, however such an approach can result in other systematic errors. For instance, in earlier experiments, ignoring stationary labels for background points (i.e. no motion) results in a large increase in mean $L_2$ error in background points from 0.03 m/s to 0.40 m/s. Hence, such heuristics are only partial solutions to this problem and new ideas are warranted for approaching this dataset. We suspect that many opportunities exist for applying semi-supervised learning techniques for generalizing to unlabeled objects and leave this to future work \cite{papandreou2015weakly,Chen2020Leveraging}.

\section{Discussion}
\label{sec:discussion}
In this work we presented a new large-scale scene flow  dataset measured from LiDAR in autonomous vehicles. Specifically, by leveraging the supervised tracking labels from the \waymo, we bootstrapped a motion vector annotation for each LiDAR return.
The resulting dataset is $\sim1000\times$ larger than previous real world scene flow datasets. We also propose a series of metrics for evaluating the resulting scene flow with breakdowns based on criteria that are relevant for deploying in the real world.

Finally, we demonstrated a scalable baseline model trained on this dataset that achieves reasonable predictive performance and may be deployed for real time operation. Interestingly, our setup opens opportunities for self-supervised and semi-supervised methods \cite{papandreou2015weakly,Chen2020Leveraging,pontes2020scene}.
We hope that this dataset may provide a useful baseline for exploring such techniques and developing generic methods for scene flow estimation in AV's in the future.

\ificrafinal
\section*{Acknowledgements}
\noindent We thank Vijay Vasudevan, Benjamin Caine, Jiquan Ngiam, Brandon Yang, Pei Sun, Yuning Chai, Charles Qi, Dragomir Anguelov, Congcong Li, Jiyang Gao, James Guo, and Yin Zhou for their comments and suggestions. Additionally, we thank the larger Google Brain and Waymo Perception teams for their support.
\fi

\bibliographystyle{IEEEtran}
\bibliography{paper}

\begin{thebibliography}{10}
\providecommand{\url}[1]{#1}
\csname url@rmstyle\endcsname
\providecommand{\newblock}{\relax}
\providecommand{\bibinfo}[2]{#2}
\providecommand\BIBentrySTDinterwordspacing{\spaceskip=0pt\relax}
\providecommand\BIBentryALTinterwordstretchfactor{4}
\providecommand\BIBentryALTinterwordspacing{\spaceskip=\fontdimen2\font plus
\BIBentryALTinterwordstretchfactor\fontdimen3\font minus
  \fontdimen4\font\relax}
\providecommand\BIBforeignlanguage[2]{{%
\expandafter\ifx\csname l@#1\endcsname\relax
\typeout{** WARNING: IEEEtran.bst: No hyphenation pattern has been}%
\typeout{** loaded for the language `#1'. Using the pattern for}%
\typeout{** the default language instead.}%
\else
\language=\csname l@#1\endcsname
\fi
#2}}

\bibitem{forsyth2002computer}
D.~A. Forsyth \emph{et~al.}, \emph{Computer vision: a modern approach}.\hskip
  1em plus 0.5em minus 0.4em\relax Prentice Hall Professional Technical
  Reference, 2002.

\bibitem{thrun2006stanley}
S.~Thrun \emph{et~al.}, ``Stanley: The robot that won the darpa grand
  challenge,'' \emph{Journal of field Robotics}, vol.~23, pp. 661--692, 2006.

\bibitem{casas2018intentnet}
S.~Casas \emph{et~al.}, ``Intentnet: Learning to predict intention from raw
  sensor data,'' in \emph{{CoRL}}, 2018, pp. 947--956.

\bibitem{chai2019multipath}
Y.~Chai \emph{et~al.}, ``Multipath: Multiple probabilistic anchor trajectory
  hypotheses for behavior prediction,'' in \emph{{CoRL}}, 2019.

\bibitem{luo2018fast}
W.~Luo \emph{et~al.}, ``Fast and furious: Real time end-to-end 3d detection,
  tracking and motion forecasting with a single convolutional net,'' in
  \emph{{CVPR}}, 2018, pp. 3569--3577.

\bibitem{mahjourian2018unsupervised}
R.~Mahjourian \emph{et~al.}, ``Unsupervised learning of depth and ego-motion
  from monocular video using 3d geometric constraints,'' in \emph{{CVPR}},
  2018.

\bibitem{liu2019flownet3d}
X.~Liu \emph{et~al.}, ``Flownet3d: Learning scene flow in 3d point clouds,'' in
  \emph{{CVPR}}, 2019, pp. 529--537.

\bibitem{wu2019pointpwc}
W.~Wu \emph{et~al.}, ``Pointpwc-net: A coarse-to-fine network for supervised
  and self-supervised scene flow estimation on 3d point clouds,'' \emph{arXiv
  preprint arXiv:1911.12408}, 2019.

\bibitem{gu2019hplflownet}
X.~Gu \emph{et~al.}, ``Hplflownet: Hierarchical permutohedral lattice flownet
  for scene flow estimation on large-scale point clouds,'' in \emph{{CVPR}},
  2019.

\bibitem{menze2015object}
M.~Menze \emph{et~al.}, ``Object scene flow for autonomous vehicles,'' in
  \emph{{CVPR}}, 2015, pp. 3061--3070.

\bibitem{geiger2012we}
A.~Geiger \emph{et~al.}, ``Are we ready for autonomous driving? the kitti
  vision benchmark suite,'' in \emph{{CVPR}}, 2012, pp. 3354--3361.

\bibitem{mayer2016flyingthings3d}
N.~Mayer \emph{et~al.}, ``A large dataset to train convolutional networks for
  disparity, optical flow, and scene flow estimation,'' in \emph{{CVPR}}, 2016.

\bibitem{wang2020flownet3d++}
Z.~Wang \emph{et~al.}, ``Flownet3d++: Geometric losses for deep scene flow
  estimation,'' in \emph{The IEEE Winter Conference on Applications of Computer
  Vision}, 2020, pp. 91--98.

\bibitem{liu2019meteornet}
X.~Liu \emph{et~al.}, ``Meteornet: Deep learning on dynamic 3d point cloud
  sequences,'' in \emph{{CVPR}}, 2019, pp. 9246--9255.

\bibitem{sun2020scalability}
P.~Sun \emph{et~al.}, ``Scalability in perception for autonomous driving: Waymo
  open dataset,'' in \emph{{CVPR}}, 2020, pp. 2446--2454.

\bibitem{saxena2006learning}
A.~Saxena \emph{et~al.}, ``Learning depth from single monocular images,'' in
  \emph{{NeurIPS}}, 2006, pp. 1161--1168.

\bibitem{scharstein2002taxonomy}
D.~Scharstein \emph{et~al.}, ``A taxonomy and evaluation of dense two-frame
  stereo correspondence algorithms,'' \emph{{IJCV}}, vol.~47, pp. 7--42, 2002.

\bibitem{pfeiffer2013exploiting}
D.~Pfeiffer \emph{et~al.}, ``Exploiting the power of stereo confidences,'' in
  \emph{{CVPR}}, 2013, pp. 297--304.

\bibitem{baker2011database}
S.~Baker \emph{et~al.}, ``A database and evaluation methodology for optical
  flow,'' \emph{{IJCV}}, vol.~92, no.~1, pp. 1--31, 2011.

\bibitem{kondermann2012performance}
D.~Kondermann \emph{et~al.}, ``On performance analysis of optical flow
  algorithms,'' in \emph{Outdoor and Large-Scale Real-World Scene
  Analysis}.\hskip 1em plus 0.5em minus 0.4em\relax Springer, 2012, pp.
  329--355.

\bibitem{morales2010ground}
S.~Morales \emph{et~al.}, ``Ground truth evaluation of stereo algorithms for
  real world applications,'' in \emph{{ACCV}}.\hskip 1em plus 0.5em minus
  0.4em\relax Springer, 2010, pp. 152--162.

\bibitem{ladicky2012joint}
L.~Ladick{\`y} \emph{et~al.}, ``Joint optimization for object class
  segmentation and dense stereo reconstruction,'' \emph{{IJCV}}, vol. 100, pp.
  122--133, 2012.

\bibitem{butler2012naturalistic}
D.~J. Butler \emph{et~al.}, ``A naturalistic open source movie for optical flow
  evaluation,'' in \emph{{ECCV}}.\hskip 1em plus 0.5em minus 0.4em\relax
  Springer, 2012, pp. 611--625.

\bibitem{wang2018deep}
S.~Wang \emph{et~al.}, ``Deep parametric continuous convolutional neural
  networks,'' in \emph{{CVPR}}, 2018, pp. 2589--2597.

\bibitem{lee2020pillarflow}
K.-H. Lee \emph{et~al.}, ``Pillarflow: End-to-end birds-eye-view flow
  estimation for autonomous driving,'' \emph{IROS}, 2020.

\bibitem{pontes2020scene}
J.~Pontes \emph{et~al.}, ``Scene flow from point clouds with or without
  learning,'' \emph{International Conference on 3D Vision}, 2020.

\bibitem{chang2019argoverse}
M.-F. Chang \emph{et~al.}, ``Argoverse: 3d tracking and forecasting with rich
  maps,'' in \emph{{CVPR}}, 2019, pp. 8748--8757.

\bibitem{caesar2020nuscenes}
H.~Caesar \emph{et~al.}, ``nuscenes: A multimodal dataset for autonomous
  driving,'' in \emph{{CVPR}}, 2020, pp. 11\,621--11\,631.

\bibitem{houston2020one}
J.~Houston \emph{et~al.}, ``One thousand and one hours: Self-driving motion
  prediction dataset,'' \emph{arXiv preprint arXiv:2006.14480}, 2020.

\bibitem{behl2019pointflownet}
A.~Behl \emph{et~al.}, ``Pointflownet: Learning representations for rigid
  motion estimation from point clouds,'' in \emph{{CVPR}}, 2019, pp.
  7962--7971.

\bibitem{fan2019pointrnn}
H.~Fan \emph{et~al.}, ``Pointrnn: Point recurrent neural network for moving
  point cloud processing,'' \emph{arXiv preprint arXiv:1910.08287}, 2019.

\bibitem{wu2020motionnet}
P.~Wu \emph{et~al.}, ``Motionnet: Joint perception and motion prediction for
  autonomous driving based on bird's eye view maps,'' in \emph{{CVPR}}, 2020.

\bibitem{dewan2016rigid}
A.~Dewan \emph{et~al.}, ``Rigid scene flow for 3d lidar scans,'' in
  \emph{{IROS}}, 2016.

\bibitem{ushani2017learning}
A.~Ushani \emph{et~al.}, ``A learning approach for real-time temporal scene
  flow estimation from lidar data,'' in \emph{{ICRA}}, 2017, pp. 5666--5673.

\bibitem{ushani2018feature}
A.~K. Ushani \emph{et~al.}, ``Feature learning for scene flow estimation from
  lidar,'' in \emph{{CoRL}}, 2018, pp. 283--292.

\bibitem{bousmalis2018using}
K.~Bousmalis \emph{et~al.}, ``Using simulation and domain adaptation to improve
  efficiency of deep robotic grasping,'' in \emph{{ICRA}}, 2018.

\bibitem{saxena2008robotic}
A.~Saxena \emph{et~al.}, ``Robotic grasping of novel objects using vision,''
  \emph{The Int'l Journal of Robotics Research}, vol.~27, pp. 157--173, 2008.

\bibitem{viereck2017learning}
U.~Viereck \emph{et~al.}, ``Learning a visuomotor controller for real world
  robotic grasping using simulated depth images,'' \emph{arXiv preprint
  arXiv:1706.04652}, 2017.

\bibitem{gualtieri2016high}
M.~Gualtieri \emph{et~al.}, ``High precision grasp pose detection in dense
  clutter,'' in \emph{{IROS}}, 2016, pp. 598--605.

\bibitem{yu2020bdd100k}
F.~Yu \emph{et~al.}, ``Bdd100k: A diverse driving dataset for heterogeneous
  multitask learning,'' in \emph{{CVPR}}, 2020, pp. 2636--2645.

\bibitem{qi2017pointnet}
C.~R. Qi \emph{et~al.}, ``Pointnet: Deep learning on point sets for 3d
  classification and segmentation,'' in \emph{{CVPR}}, 2017, pp. 652--660.

\bibitem{zhou2019multiview}
Y.~Zhou \emph{et~al.}, ``End-to-end multi-view fusion for 3d object detection
  in lidar point clouds,'' in \emph{{CoRL}}, 2019.

\bibitem{ronneberger2015unet}
O.~Ronneberger \emph{et~al.}, ``U-net: Convolutional networks for biomedical
  image segmentation,'' in \emph{{MICCAI}}.\hskip 1em plus 0.5em minus
  0.4em\relax Springer, 2015, pp. 234--241.

\bibitem{lang2018pointpillars}
A.~H. Lang \emph{et~al.}, ``Pointpillars: Fast encoders for object detection
  from point clouds,'' \emph{arXiv preprint arXiv:1812.05784}, 2018.

\bibitem{jund2021scalable}
P.~Jund \emph{et~al.}, ``Scalable scene flow from point clouds in the real
  world,'' \emph{arXiv preprint arXiv:2103.01306}, 2021.

\bibitem{zhou2008real}
K.~Zhou \emph{et~al.}, ``Real-time kd-tree construction on graphics hardware,''
  \emph{ACM Transactions on Graphics (TOG)}, vol.~27, no.~5, pp. 1--11, 2008.

\bibitem{chen2019fast}
Y.~Chen \emph{et~al.}, ``Fast neighbor search by using revised kd tree,''
  \emph{Information Sciences}, vol. 472, pp. 145--162, 2019.

\bibitem{odena2016deconvolution}
A.~Odena \emph{et~al.}, ``Deconvolution and checkerboard artifacts,''
  \emph{Distill}, 2016.

\bibitem{ding20201st}
Z.~Ding \emph{et~al.}, ``1st place solution for waymo open dataset
  challenge--3d detection and domain adaptation,'' \emph{arXiv preprint
  arXiv:2006.15505}, 2020.

\bibitem{mcnaughton2011motion}
M.~McNaughton \emph{et~al.}, ``Motion planning for autonomous driving with a
  conformal spatiotemporal lattice,'' in \emph{{ICRA}}, 2011, pp. 4889--4895.

\bibitem{chu2012local}
K.~Chu \emph{et~al.}, ``Local path planning for off-road autonomous driving
  with avoidance of static obstacles,'' \emph{IEEE Transactions on Intelligent
  Transportation Systems}, vol.~13, no.~4, pp. 1599--1616, 2012.

\bibitem{dolgov2008practical}
D.~Dolgov \emph{et~al.}, ``Practical search techniques in path planning for
  autonomous driving,'' in \emph{{AAAI}}, vol. 1001, 2008, pp. 18--80.

\bibitem{papandreou2015weakly}
G.~Papandreou \emph{et~al.}, ``Weakly-and semi-supervised learning of a deep
  convolutional net for semantic image segmentation,'' in \emph{{ICCV}}, 2015.

\bibitem{Chen2020Leveraging}
L.-C. Chen \emph{et~al.}, ``Leveraging semi-supervised learning in video
  sequences for urban scene segmentation.'' in \emph{{ECCV}}, 2020.

\bibitem{filatov2020any}
A.~Filatov, A.~Rykov, and V.~Murashkin, ``Any motion detector: Learning
  class-agnostic scene dynamics from a sequence of lidar point clouds,''
  \emph{arXiv preprint arXiv:2004.11647}, 2020.

\bibitem{kingma2014adam}
D.~P. Kingma and J.~Ba, ``Adam: A method for stochastic optimization,''
  \emph{arXiv preprint arXiv:1412.6980}, 2014.

\bibitem{shen2019lingvo}
J.~Shen, P.~Nguyen, Y.~Wu, Z.~Chen, M.~X. Chen, Y.~Jia, A.~Kannan, T.~Sainath,
  Y.~Cao, C.-C. Chiu, \emph{et~al.}, ``Lingvo: a modular and scalable framework
  for sequence-to-sequence modeling,'' \emph{arXiv preprint arXiv:1902.08295},
  2019.

\bibitem{ngiam2019starnet}
J.~Ngiam, B.~Caine, W.~Han, B.~Yang, Y.~Chai, P.~Sun, Y.~Zhou, X.~Yi,
  O.~Alsharif, P.~Nguyen, \emph{et~al.}, ``Starnet: Targeted computation for
  object detection in point clouds,'' \emph{arXiv preprint arXiv:1908.11069},
  2019.

\bibitem{glorot2010understanding}
X.~Glorot and Y.~Bengio, ``Understanding the difficulty of training deep
  feedforward neural networks,'' in \emph{{AISTATS}}, 2010, pp. 249--256.

\end{thebibliography}

\ificrafinal
\clearpage
\appendix

\section*{Bootstrapping Ground Truth Annotations}
\label{section:ground-truth-details}

\noindent In this section, we discuss in detail several practical considerations for the method for computing scene flow annotations (\secref{sec:computing_groundtruth}).
One challenge in this context is the lack of correspondence between the observed points in $\pointcloudAtT{-1}$ and $\pointcloudAtT{0}$.
In our work, we choose to make flow predictions (and thus compute annotations) for the points at the current time step, $\pointcloudAtT{0}$. As opposed to doing so for $\pointcloudAtT{-1}$, we believe that explicitly assigning flow predictions to the points in the most recent frame is advantageous to an AV that needs to reason about and react to the environment in real time. Additionally, the motion between $\pointcloudAtT{-1}$ and $\pointcloudAtT{0}$ is a reasonable approximation for the flow at $\tcurr$ when considering a high \lidar acquisition frame rate and assuming a constant velocity between consecutive frames.
\linebreak\linebreak
\noindent\textbf{Calculating the transformation for the motion of an object.} The goal of this section is to describe the calculation of $\transformAtT{\Delta}$ used to transform a point $\point_{0}$ to its corresponding position at $\tprev$ based on the motion of the labeled object to which it belongs. We leverage the 3D label boxes to circumvent the point-wise correspondence problem between $\pointcloudAtT{0}$ and $\pointcloudAtT{-1}$ and estimate the position of the points belonging to an object in $\pointcloudAtT{0}$ as they {\it would have been observed} at $\tprev$. We compute the flow vector for each point at $\tcurr$ using its displacement over the duration of $\Delta_t = \tcurr - \tprev$. Let point clouds $\pointcloudAtT{-1}$ and $\pointcloudAtT{0}$ be represented in the reference frame of the AV at their corresponding time steps. We identify the set of objects $\objects_{0}$ at $\tcurr$ based on the annotated 3D boxes of the corresponding scene. We express the pose of an object $\object$ in the AV frame as a homogeneous transformation matrix $\transform$ consisting of 3D translation and rotational components derived from the pose of tracked objects.
For each object $\object\in\objects_{0}$, we first use its pose $\transformAtT{-1}$ relative to the~\av~at $\tprev$ and compensate for ego motion to compute its pose $\transformPrimeAtT{-1}$ at $\tprev$ but with respect to the \av~frame at $\tcurr$. This is straightforward given knowledge of the poses of the~\av~at the corresponding time steps in the dataset, e.g. from a localization system. Accordingly, we compute the rigid body transform $\transformAtT{\Delta}$ used to transform points belonging to object $\object$ at time $\tcurr$ to their corresponding position at $\tprev$, i.e. $\transformAtT{\Delta}~\coloneqq~\transformPrimeAtT{-1} \cdot \transformAtT{0}^{-1}$. %
\linebreak\linebreak
\noindent\textbf{Rigid body assumption.} Our approach for scene flow annotation assumes the 3D label boxes correspond to rigid bodies, allowing us to compute the point-wise correspondences between two frames. Although this is a common assumption in the literature (especially for labeled vehicles \cite{menze2015object}), this does not necessarily apply to non-rigid objects such as pedestrians. However, we found this to be a reasonable approximation in our work on the \waymo~for two reasons. First, we derive our annotations from frames measured at high frequency (i.e. 10 Hz) such that object deformations are minimal between adjacent frames. Second, the number of observed points on objects like pedestrians is typically small making any deviations from a rigid assumption to be of statistically minimal consequence.
\linebreak\linebreak
\noindent\textbf{Objects with no matching previous frame labels.} In some cases, an object $\object\in\objects_{\tcurr}$ with a label box at $\tcurr$ will not have a corresponding label at $\tprev$, e.g. the object is first observable at $\tcurr$. Without information about the motion of the object between $\tprev$ and $\tcurr$, we choose to annotate its points as having invalid flow. While we can still use them to encode the scene and extract features during model training, this annotation allows us to exclude them from model weight updates and scene flow evaluation metrics.
\linebreak\linebreak
\noindent\textbf{Background points.} Since typically most of the world is stationary (e.g. buildings, ground, vegetation), it is important to reflect this in the dataset. Having compensated for ego motion, we assign zero motion for all unlabeled points in the scene, and additionally annotate them as belonging to the ``background'' class. Although this holds for the vast majority of unlabeled points, there will always exist rare moving objects in the scene that were not manually annotated with label boxes (e.g. animals). In the absence of label boxes, points of such objects will receive a stationary annotation by default. Nonetheless, we recognize the importance of enabling a model to predict motion on unlabeled objects, as it is crucial for an~\av~to safely react to rare, moving objects. In~\secref{sec:generalization}, we highlight this challenge and discuss opportunities for employing this dataset as a benchmark for semi-supervised and self-supervised learning.
\linebreak\linebreak
\noindent\textbf{Coordinate frame of reference.} As opposed to most other works \cite{geiger2012we,menze2015object}, we account for ego motion in our scene flow annotations. Not only does this better reflect the fact that most of the world is stationary, but it also improves the interpretability of flow annotations, predictions, and evaluation metrics. In addition to compensating for ego motion when computing flow annotations at $\tcurr$, we also transform $\pointcloudAtT{{-1}}$, the scene at $\tprev$, to the reference frame of the~\av~at $\tcurr$ when learning and inferring scene flow. We argue that this is more realistic for~\av~applications in which ego motion is available from IMU/GPS sensors \cite{thrun2006stanley}. Furthermore, having a consistent coordinate frame for both input frames lessens the burden on a model to correspond moving objects between frames \cite{filatov2020any} as explored in~\appref{sec:ego-motion}.

\section*{Model Architecture and Training Details}
\label{section:model-architecture}

Figure \ref{fig:model-diagram} provides an overview of \modelname. We discuss each section of the architecture in turn and provide additional parameters in Table \ref{tab:architecture}.

The model architecture contains in total 5,233,571 parameters. The vast majority of the parameters (4,212,736) reside in the standard convolution architecture \cite{lang2018pointpillars}. An additional large set of parameters (1,015,808) reside in later layers that perform upsampling with a skip connection \cite{odena2016deconvolution}. Finally, a small number of parameters (544) are dedicated to featurizing each point cloud point \cite{qi2017pointnet} as well as performing the final regression on to the motion flow (4483). Note that both of these latter sets of parameters are purposefully small because they are applied to all $N$ points in the LiDAR point cloud.

\modelname~uses a top-down U-Net to process the pillarized features. Consequently, the model can only predict flow for points inside the pillar grid. Points outside the $x$-$y$ dimensions of the grid or outside the $z$ dimension bounds for the pillars are marked as invalid and receive no predictions. To extend the scope of the grid, one can either make the pillar size larger or increase the size of the pillar grid. In our work, we use a $170\times170$~m grid (centered at the AV) represented by $512\times512$ pillars ($\sim$ $0.33\times0.33$~m pillars). For the $z$ dimension, we consider the valid pillar range to be from $-3$ m to $+3$ m.

The model was trained for 19 epochs on the \waymo~training set using the Adam optimizer \cite{kingma2014adam}. The model was written in Lingvo \cite{shen2019lingvo} and forked from the open-source repository version of PointPillars 3D object detection \cite{lang2018pointpillars,ngiam2019starnet} \footnote{\,\url{https://github.com/tensorflow/lingvo/}}. The training set contains a label imbalance, vastly over-representing background stationary points. In early experiments, we explored a hyper-parameter to artificially downweight background points and found that weighing down the $L_2$ loss by a factor of 0.1 provided good performance.

\section*{Compensating for Ego Motion}
\label{sec:ego-motion}

In~\secref{sec:computing_groundtruth}, we argue that compensating for ego motion in the scene flow annotations improves the interpretability of flow predictions and highlights important patterns and biases in the dataset, e.g. slow vs fast objects. When training our proposed \modelname~model, we also compensate for ego motion by transforming both \lidar frames to the reference frame of the \av~at $\tcurr$, the time step at which we predict flow. This is convenient in practice given that ego motion information is easily available from the localization module of an \av. We hypothesize that this lessens the burden on the model, because the model does not have to implicitly learn to compensate for the motion of the \av.

We validate this hypothesis in a preliminary experiment where we compare the performance of the model reported in~\secref{sec:results} to a model trained on the same dataset but without compensating for ego motion in the input point clouds. Consequently, this model has to implicitly learn how to compensate for ego motion.
\tabref{table:ego-motion-ablation} shows the mean $L_2$ error for two such models. We observe that mean $L_2$ error increases substantially when ego motion is not compensated for across all object types and across moving and stationary objects. This is also consistent with previous works \cite{filatov2020any}. We also ran a similar experiment where the model consumes non ego motion compensated point clouds, but instead subtracts ego motion from the predicted flow during training and evaluation. We found slightly better performance for moving objects for this setup, but the performance is still far short of the performance achieved when compensating for ego motion directly in the input. Further research is needed to effectively learn a model that can implicitly account for ego motion.

\section*{Measurements of Latency}
\label{sec:latency}

In this section we provide additional details for how the latency numbers for Table \ref{table:latency-vs-points} were calculated. All calculations were performed on a standard NVIDIA Tesla P100 GPU with a batch size of 1. The latency is averaged over 90 forward passes, excluding 10 warm up runs. Latency for the baseline models, HPLFlowNet \cite{gu2019hplflownet} and FlowNet3D \cite{liu2019flownet3d,wang2020flownet3d++} included any preprocessing necessary to perform inference.
For HLPFlowNet and FlowNet3D, we used the implementations provided by the authors and did not alter hyperparameters. Note that this is in favor of these models, as they were tuned  for point clouds covering a much smaller area compared to the Waymo Open Dataset.

\begin{table*}[hb]
\small
    \centering
    \rowcolors{3}{gray!15}{white}
     \bgroup
     \def\arraystretch{1.2}%
    \begin{tabular}{c|ccc|ccc|ccc|c}
        \toprule
        ego motion &
        \multicolumn{3}{c|}{vehicle} & \multicolumn{3}{c|}{pedestrian} & \multicolumn{3}{c|}{cyclist} & \multirow{2}{*}{background} \\
        compensated for& all & moving & stationary & all & moving & stationary & all & moving & stationary & \\
        \midrule
        yes & 0.18 & 0.54 & 0.05   &   0.25 & 0.32 & 0.10 & 0.51 & 0.57 & 0.10 & 0.07 \\
        no & 0.36 & 1.16 & 0.08 & 0.49 & 0.63 &  0.17 & 1.21 & 1.34 & 0.14 & 0.07 \\
        \bottomrule
    \end{tabular}
    \egroup
    \caption{\textbf{Accounting for ego motion in the input significantly improves performance}. All reported values correspond to the mean $L_2$ error in m/s. The first column refers to whether or not we transform the point cloud at $\tprev$ to the reference frame of the~\av~at $\tcurr$. For both models, we evaluate the error in predicting scene flow annotations as described in~\secref{sec:computing_groundtruth}, i.e. with ego motion compensated for.}
    \label{table:ego-motion-ablation}
\end{table*}

\section*{Dataset Format for Annotations}
\label{sec:appendix-format}

In order to access the data, please go to  \url{http://www.waymo.com/open} and click on {\tt Access Waymo Open Dataset}, which requires a user to sign in with Google and accept the Waymo Open Dataset license terms. After logged in, please visit \url{https://pantheon.corp.google.com/storage/browser/waymo_open_dataset_scene_flow} to download the labels.

We extend the \waymo~to include the scene flow labels for the training and validation datasets splits. For each LiDAR, we add a new range image through the field \texttt{range\_image\_flow\_compressed} in the message \texttt{dataset.proto:RangeImage}.
The range image is a 3D tensor of shape $[H, W, 4]$ where $H$ and $W$ are the height and width of the \lidar~scan. For the \lidar returns at point $(i, j)$, we provide annotations in the range image where $[i, j, 0$:$3]$ corresponds to the estimated velocity components for the return along $x, y$ and $z$ axes, respectively. Finally, the value stored in the range image at $[i, j, 3]$ contains an integer class label.

\begin{table*}[h]
\small
\centering
\bgroup
\def\arraystretch{1.2}%
\begin{tabular}{lccccccccc}
\toprule
Meta-Arch & Name & Input(s) & Operation & Kernel & Stride & BN? & Output Size & Depth & \# Param \\
\midrule
{\bf Architecture}  \\
& A & $N$ pts & MLP & -- & -- & Yes &  $N$ pts & 64 & 544 \\
& B & A & Snap-To-Grid & -- & -- & -- &  $512\times512$ & 64 & 0 \\
\\
& C & B & Convolution & $3\times3$ & 2 & Yes &  $256\times256$ & 64 & 37120 \\
& D & C & Convolution & $3\times3$ & 1 & Yes & $256\times256$ & 64 & 37120 \\
& E & D & Convolution & $3\times3$ & 1 & Yes & $256\times256$ & 64 & 37120 \\
& F & E & Convolution & $3\times3$ & 1 & Yes & $256\times256$ & 64 & 37120 \\

& G & F & Convolution & $3\times3$ & 2 & Yes & $128\times128$ & 128 & 74240 \\
& H & G & Convolution & $3\times3$ & 1 & Yes & $128\times128$ & 128 & 147968 \\
& I & H & Convolution & $3\times3$ & 1 & Yes & $128\times128$ & 128 & 147968 \\
& J & I & Convolution & $3\times3$ & 1 & Yes & $128\times128$ & 128 & 147968 \\
& K & J & Convolution & $3\times3$ & 1 & Yes & $128\times128$ & 128 & 147968 \\
& L & K & Convolution & $3\times3$ & 1 & Yes & $128\times128$ & 128 & 147968 \\

& M & L & Convolution & $3\times3$ & 2 & Yes & $64\times64$ & 256 & 295936 \\
& N & M & Convolution & $3\times3$ & 1 & Yes & $64\times64$ & 256 & 590848 \\
& O & N & Convolution & $3\times3$ & 1 & Yes & $64\times64$ & 256 & 590848 \\
& P & O & Convolution & $3\times3$ & 1 & Yes & $64\times64$ & 256 & 590848 \\
& Q & P & Convolution & $3\times3$ & 1 & Yes & $64\times64$ & 256 & 590848 \\
& R & Q & Convolution & $3\times3$ & 1 & Yes & $64\times64$ & 256 & 590848 \\

& S & R$^*$, L$^*$, 128, 128 & Upsample-Skip & -- & -- & No & $128\times128$ & 128 & 540672 \\

& T & S, F$^*$, 128, 64 & Upsample-Skip & -- & -- & No & $256\times256$ & 128 & 311296 \\

& U & T, B$^*$, 64, 64 & Upsample-Skip & -- & -- & No & $512\times512$ & 64 & 126976 \\

& V & U & Convolution & $3\times3$ & 1 & No & $512\times512$ & 64 & 36864 \\
\\
& W & V & Ungrid & -- & -- & -- &  $N$ pts & 64 & 0 \\
& X & W, B & Concatenate & -- & -- & -- &  $N$ pts & 128 & 0 \\
& Y & X & Linear & -- & -- & No & $N$ pts & 32 & 4384 \\
& Z & Y & Linear & -- & -- & No & $N$ pts & 3 & 99 \\

\midrule

\multicolumn{2}{l}{{\bf Upsample-Skip} $(\alpha, \beta, \mathsf{d}, \mathsf{d_b})$} \\
& U1 & $\alpha$ & Convolution & $1\times1$ & 1 & No & $h_\alpha\times w_\alpha$ & $\mathsf{d_b}$ &\\
& U2 & U1 & Bilinear Interp. & -- & $\frac{1}{2}$ & No & $h_\beta\times w_\beta$ & $\mathsf{d_b}$ & \\
& U3 & $\beta$ & Convolution & $1\times1$ & 1 & No & $h_\beta\times w_\beta$ & $\mathsf{d_b}$ & \\
& U4 & U2, U3 & Concatenate & -- & -- & No & $h_\beta\times w_\beta$ & 2$\mathsf{d_b}$ & \\
& U5 & U4 & Convolution & $3\times3$ & 1 & No & $h_\beta\times w_\beta$ & $\mathsf{d}$ & \\
& U6 & U5 & Convolution & $3\times3$ & 1 & No & $h_\beta\times w_\beta$ & $\mathsf{d}$ & \\

\midrule

\multicolumn{2}{l}{Padding mode} & \multicolumn{4}{l}{SAME} \\
\multicolumn{2}{l}{Optimizer} & \multicolumn{4}{l}{Adam \cite{kingma2014adam} ($lr = 1\mathrm{e}{-6}$, $\beta_1 = 0.9$, $\beta_2 = 0.999$)}  \\
\multicolumn{2}{l}{Batch size} & \multicolumn{4}{l}{64} \\
\multicolumn{2}{l}{Weight initialization}  & \multicolumn{4}{l}{Xavier-Glorot \cite{glorot2010understanding}}  \\
\multicolumn{2}{l}{Weight decay}  & \multicolumn{4}{l}{None} \\

\bottomrule

\end{tabular}
\egroup
\vspace{0.2cm}
\caption{\label{tab:architecture} {\bf \modelname\,architecture and training details.} The network receives as input $N$ 3D points with additional point cloud features and outputs $N$ 3D points corresponding to the predicted 3D motion vector of each point. This output corresponds to the output of layer $Z$. All layers employ a ReLU nonlinearity except for layers $S-Z$ which employ no nonlinearity. The shape of tensor $x$ is denoted $h_x\times w_x$. The Upsample-Skip layer receives two tensors $\alpha$ and $\beta$ and a scalar depth $\mathsf{d}$ as input and outputs tensor $U6$. Inputs with $^*$ denote the concatenation of the respective layers of the weight-sharing encoders.}
\end{table*}

\fi

\end{document}